\documentclass[10pt,twocolumn,letterpaper]{article}
\usepackage{iccv}

\usepackage{amsmath}
\usepackage{amsfonts}
\usepackage{amssymb}
\usepackage{amsthm}
\usepackage{bm}
\usepackage{bbm}
\usepackage{mathtools}
\usepackage{enumitem}
\usepackage{thmtools,thm-restate}
\usepackage[ruled,vlined,linesnumbered]{algorithm2e}
\SetAlFnt{\small}
\usepackage{color}
\usepackage[dvipsnames]{xcolor}
\usepackage{graphicx}
\usepackage{comment}

\theoremstyle{plain}

\theoremstyle{definition}

\theoremstyle{remark}

\def\bomega{{\bm\omega}}

\def\btheta{{\bm\theta}}

\def\CC{\mathcal{C}}
\def\DD{\mathcal{D}}\def\FF{\mathcal{F}}

\def\LL{\mathcal{L}}
\def\NN{\mathcal{N}}\def\OO{\mathcal{O}}

\def\Ib{\mathbf{I}}

\def\eb{\mathbf{e}}

\def\ob{\mathbf{o}}
\def\rb{\mathbf{r}}

\def\vb{\mathbf{v}}\def\wb{\mathbf{w}}\def\xb{\mathbf{x}}
\def\yb{\mathbf{y}}\def\zb{\mathbf{z}}

\def\Ebb{\mathbb{E}}

\def\Rbb{\mathbb{R}}
\def\Sbb{\mathbb{S}}

\def\R{\Rbb}

\def\*{\star}

\newcommand{\norm}[1]{ \| #1 \|  }

\DeclareMathOperator*{\argmax}{arg\,max}

\newcommand{\E}{\Ebb}

\usepackage{etex,etoolbox}
\usepackage{environ}

\makeatletter
\providecommand{\@fourthoffour}[4]{#4}
\newcommand\fixstatement[2][\proofname\space of]{%
	\ifcsname thmt@original@#2\endcsname
	\AtEndEnvironment{#2}{%
		\xdef\pat@label{\expandafter\expandafter\expandafter
			\@fourthoffour\csname thmt@original@#2\endcsname\space\@currentlabel}%
		\xdef\pat@proofof{\@nameuse{pat@proofof@#2}}%
	}%
	\else
	\AtEndEnvironment{#2}{%
		\xdef\pat@label{\expandafter\expandafter\expandafter
			\@fourthoffour\csname #1\endcsname\space\@currentlabel}%
		\xdef\pat@proofof{\@nameuse{pat@proofof@#2}}%
	}%
	\fi
	\@namedef{pat@proofof@#2}{#1}%
}

\newcounter{proofcount}

\NewEnviron{proofatend}{%
	\edef\next{%
		\noexpand\begin{proof}[\pat@proofof\space\pat@label]%
			\unexpanded\expandafter{\BODY}}%
		\global\toks\numexpr\prooftoks+\value{proofcount}\relax=\expandafter{\next\end{proof}}
	\stepcounter{proofcount}}

\def\printproofs{%
	\count@=\z@
	\loop
	\the\toks\numexpr\prooftoks+\count@\relax
	\ifnum\count@<\value{proofcount}%
	\advance\count@\@ne
	\repeat}
\makeatother

\fixstatement{lemma}
\fixstatement{theorem}
\fixstatement{proposition}
\fixstatement{corollary}
\usepackage{times}
\usepackage{epsfig}
\usepackage{graphicx}
\usepackage{amsmath}
\usepackage{amssymb}
\usepackage{mathtools}
\usepackage{multirow}
\usepackage{booktabs}
\usepackage{colortbl}
\usepackage{hhline}
\usepackage{xcolor}
\usepackage{pstricks}
\usepackage[toc,page]{appendix}

\usepackage[pagebackref=true,breaklinks=true,colorlinks,bookmarks=false]{hyperref}
\usepackage[capitalise]{cleveref}

\iccvfinalcopy 

\def\iccvPaperID{6665} 

\ificcvfinal\fi

\usepackage[ruled,linesnumbered]{algorithm2e}
\usepackage{etoolbox}
\makeatletter
\patchcmd{\@algocf@start}
  {-1.5em}
  {0pt}
  {}{}
\makeatother
\newcommand{\cmnt}[1]{}

\begin{document}

\title{Few-shot Weakly-Supervised Object Detection via Directional Statistics}

 \author{Amirreza Shaban\thanks{Equal Contribution. Contact at {\tt\small ashaban@uw.edu} or {\tt\small amir.rahimi@anu.edu.au}}\\
 University of Washington\\
 \and
 Amir Rahimi$^*$\\
 ANU \& ACRV\\
 \and 
 Thalaiyasingam Ajanthan\\
 ANU \& ACRV\\
 \and 
 Byron Boots\\
 University of Washington\\
 \and 
 Richard Hartley\\
 ANU \& ACRV\\
 }

\maketitle
%
%
%
\begin{abstract}
Detecting novel objects from few examples has become an emerging topic in computer vision recently. 
However, these methods need fully annotated training images to learn new object categories which limits their applicability in real world scenarios such as field robotics.
In this work, we propose a probabilistic multiple instance learning approach for few-shot Common Object Localization (COL) and few-shot Weakly Supervised Object Detection (WSOD). 
In these tasks, only image-level labels, which are much cheaper to acquire, are available. 
We find that operating on features extracted from the last layer of a pre-trained Faster-RCNN is more effective compared to previous episodic learning based few-shot COL methods. 
Our model simultaneously learns the distribution of the novel objects and localizes them via expectation-maximization steps.
As a probabilistic model, we employ von Mises-Fisher (vMF) distribution which captures the semantic information better than Gaussian distribution when applied to the pre-trained embedding space.
When the novel objects are localized, we utilize them to learn a linear appearance model to detect novel classes in new images.
Our extensive experiments 
show that the proposed method, despite being simple, outperforms strong baselines in few-shot COL and WSOD, as well as large-scale WSOD tasks.

%
\end{abstract}
\section{Introduction}
In this paper we address the problem of $N$-way, $K$-shot Weakly Supervised Object Detection (WSOD),
and develop a method with the following capabilities.  
Suppose that we are given a set of $N\times K$ previously unseen images consisting of
$K$ images of objects from each of $N$ previously unknown (novel) classes.  
These will be called the ``support images''.
Each training image has
image-level labels, indicating which classes are present in the image. 
Typically, the number
of novel classes $N$ may be up to $20$ and the number of training images $K$ from
each class may be $5$ or $10$, but there is no requirement that the
number of images in each novel class are equal. 

Given this small number of support images, the algorithm learns to find instances 
of (possibly multiple) objects from any of the novel
classes in a query image, and will put a bounding box around all such
positive instances. 
%
\begin{figure}[t]
\includegraphics[width=\linewidth]{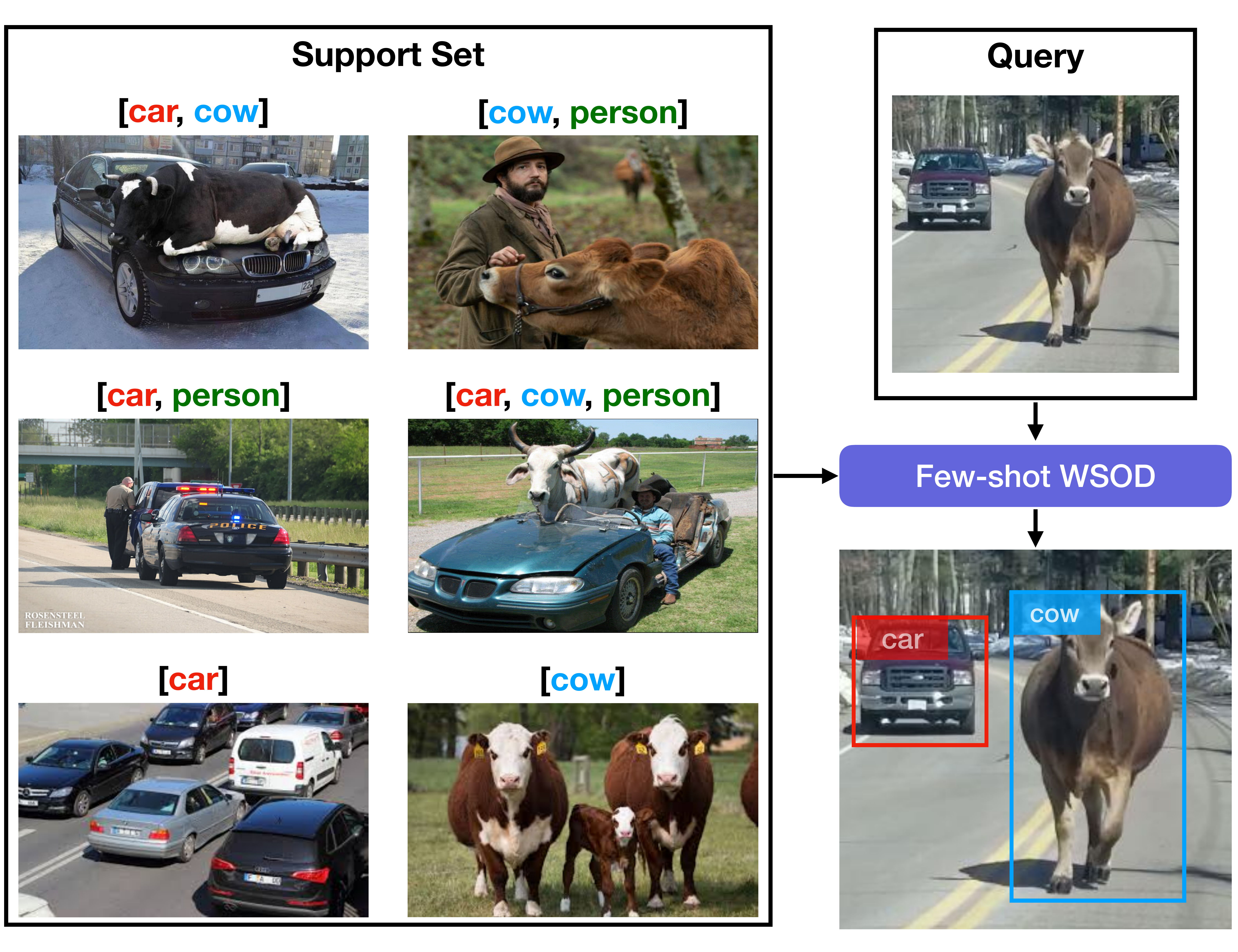}
\caption{\em\label{fig:problem_summary} Few-shot WSOD problem. Similar to the
few-shot classification problem, the input training set (support set) only
contains image labels ({\tt car}, {\tt cow} and {\tt person} are novel classes
in this example). Model learns to detect the target objects in the test (query)
image. Few-shot WSOD bridges few-shot classification and object detection 
by learning to detect the novel objects in the query images while only needs image-level labels 
for the support images.}
\vspace{-0.5cm}
\end{figure}
\begin{figure*}[t]
\begin{center}
\includegraphics[width=0.99\linewidth]{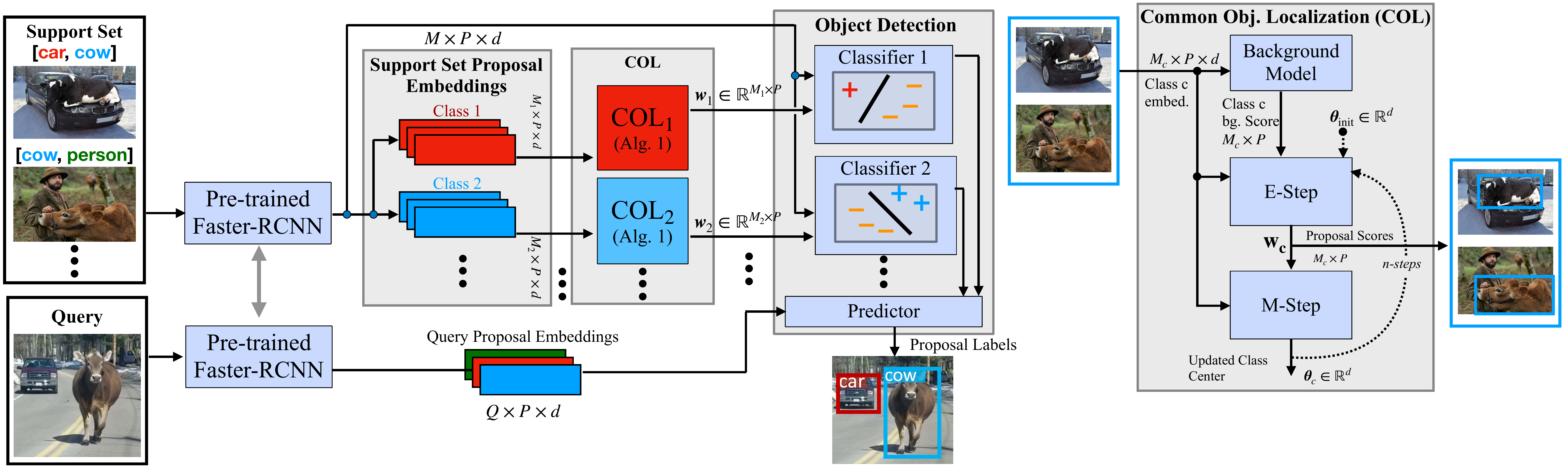}

\caption{\em \label{fig:model_summary} The feature maps are shown as the shape of their tensors. $Q$ and $M$ denote the total number of query and support images, respectively. 
A Faster-RCNN (shown in \cref{fig:faster_rcnn}) trained on the base dataset is used to extract $P$ proposals from each input image. 
The embeddings are grouped based on their corresponding image-level labels and each group is fed into a separate Common Object Localization (COL) module. 
The {\bf COL} module (shown in detail on the right) receives proposal embeddings of images of a class ($M_c$ is the number of images within class $c$) and simultaneously estimates the common class mean $\btheta_c$ and bounding-box level labels $\wb_c$ via EM steps. 
{\bf\small Object Detection} module uses the top predictions in $\wb_c$ to learn an appearance model for each novel class in the support set. This appearance model is then tested on the query proposals to detect novel objects in the query set.}

\vspace{-0.8cm}
\end{center}
\end{figure*}

As summarized in \cref{fig:problem_summary}, our system provides a flexible 
object detection algorithm that requires a very small training set of images of novel objects, where
each image is annotated only with image-level labels. 
As such, it is suitable
for classifying and detecting objects given only images provided, for
instance, by an internet image search for images of novel classes. 
In comparison to supervised few-shot object detection approaches, e.g., ~\cite{xiao2020few, wang2019meta, wang2020frustratingly, perez2020incremental}, 
where manually labeled bounding box annotations are required, this is a 
more realistic setting to learn an object detector on novel examples with applications 
like robotics~\cite{kim2020tell} or video object segmentation~\cite{lu2019see}.


We first use a Faster-RCNN network to produce bounding box proposals
of possible regions containing an object with their associated feature vectors. This network is pre-trained
on a fully annotated ``base dataset'', with bounding boxes of objects of various classes, not containing any of
the novel training classes. Then, the novel objects are learned in a two steps process: 1) A {\em common object localization} (COL) module is used first to localize the novel objects in the support images. 2) An object detection module to learn 
the novel object classes found in the COL step. We explain each of these modules in more details below.

{\bf\noindent COL module.}
To localize the novel objects in the support images, COL module finds the common object in the $K$
images provided for each of the $N$ novel classes. The COL module is 
run separately on the images from each of the $N$ novel classes.
For a novel class $c$, the input to the COL module
is the set of normalized feature vectors (corresponding to bounding boxes
provided by the proposal network) from 
the images of class $c$. 
An EM algorithm on these feature vectors
determines the direction parameter of von Mises-Fisher (vMF) distribution on the unit sphere
most likely to favour 
a common object representative from each image. The closest feature from each image
identifies the bounding box containing the common object. 
A distribution for a background class is also trained, using the base dataset to steer the 
COL away from selecting background objects.
%

{\bf\noindent Detection module. }
Once the novel objects are found by the COL module, their bounding boxes are used to train
a box classifier for each novel class $c$.
The classification is done by a 2-class (contains / does not contain
the object) classification algorithm, once again working on normalized feature vectors.
These bounding boxes (and their
associated features) are labeled either as positive
or negative for containing the object of the novel class, or not.  The positive
bounding boxes are those that are determined by the COL module
to contain the common object from class $c$, and the negative samples
are chosen from
proposals selected from the images of the other classes. 
Thus, the classifier for class $c$ is trained to 
distinguish features corresponding to bounding boxes containing an
object of class $c$ from those that do not.

Finally, at test time, a query image is passed through the proposal network,
to provide bounding boxes (and their features). These bounding boxes
are then evaluated by each of the classifiers, to determine
whether they contain a novel class object, or belong
to the background.

The proposed method is summarized in~\cref{fig:model_summary}. We make several contributions
and important observations: 1) We propose a simple yet powerful COL that uses directional statistics
for modeling. Our COL module can be built on top of off-the-shelf pre-trained Faster-RCNN models without extra parameters. 
We observe that using feature vector directions in our probabilistic model, we can 
better capture the semantic information compared to 
Gaussian distribution. To our knowledge, employing directional statistics for multiple instance learning is new.
2) We employ a detection module to extend COL to few-shot WSOD. 
To the best of our knowledge few-shot WSOD not been studied in the literature before.
3) Despite its simplicity, our method outperforms sophisticated few-shot COL
algorithms~\cite{shaban2019learning, hu2019silco} on PASCAL
VOC~\cite{pascal-voc-2007}, MS COCO~\cite{lin2014microsoft}, and ILSVRC
detection~\cite{imagenet_cvpr09} benchmarks. In WSOD, our method outperforms
recent knowledge transfer based approaches~\cite{uijlings2018revisiting, hoffman2016large} 
in both few-shot and large-scale settings.

\section{Details of Methodology}
\subsection{Few-Shot WSOD and COL Tasks Definition}
The goal in few-shot WSOD is to learn a model that, given support images $\DD_{\rm train}$ containing a set of novel classes $\LL$, detects instances of the novel classes in query images $\DD_{\rm test}$. 
The support set consists of image-label pairs $(\Ib, \yb)\in\DD_{\rm train}$ where image-level label $\yb \subseteq \LL$ is a subset of classes present in the image $\Ib$\footnote{In contrast to few-shot image classification, few-shot WSOD images can have multiple labels.}. 
The support set is typically a small $K$-shot, $N$-way set sampled from a large dataset $\DD_{\rm novel}$  with variety of novel classes $\CC_{\rm novel}$.
The sampling process for few-shot WSOD follows
rules that are similar to few-shot classification problems~\cite{lee2019meta,snell2017prototypical}. A set of $N$ classes $\LL \subset \CC_{\rm novel}$, called
target classes, are first sampled. 
Then, for each target class $c \in \LL$, $K$
images containing at least an instance of class $c$ are sampled \emph{without
replacement} to create the support set $\DD_{\rm train}$. The query set $\DD_{\rm test}$ is sampled
similarly, but unlike the support set, query labels also contain bounding box
annotations in addition to the image-level labels as the goal is to detect
target objects in the query data. These bounding box annotations are only used for evaluation. Few-shot COL~\cite{shaban2019learning, hu2019silco} is a special case of few-shot WSOD where there is
only one target class in the support set, i.e., $N=1$. 

For pre-training, the algorithm has access to a large dataset $\DD_{\rm base}$ with a set of base classes $\CC_{\rm base}$. 
Typically, there is no image in common between the base and novel
datasets. Moreover, the set of base classes is disjoint from the set of novel classes used in evaluation, i.e., $\CC_{\rm base} \cap \CC_{\rm novel} = \varnothing$.

\begin{figure}[t]
\begin{center}
\includegraphics[width=0.99\linewidth]{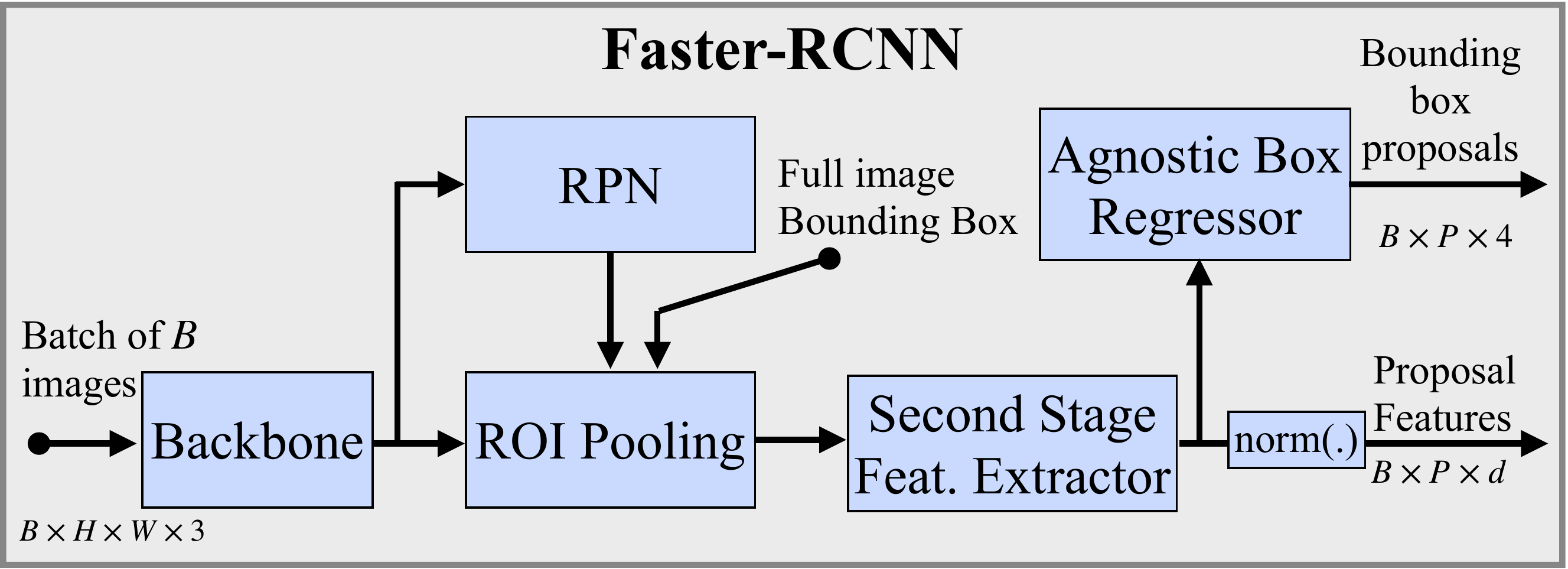}
\caption{\em \label{fig:faster_rcnn} Feature Extraction. We use a pre-trained Faster-RCNN on the base dataset to extract $P$ proposals from each input image. A $\ell_2$ normalization layer is employed to project all the features onto the unit hypersphere. }
\end{center}
\end{figure}

\subsection{Pre-training and Feature Extraction}
We pre-train a Faster-RCNN~\cite{ren2016faster} on the base dataset for bounding
box and feature extraction. The overall architecture is shown in \cref{fig:faster_rcnn}. 
To train the network, we use the original bounding
box labels within the base dataset to define the Region Proposal Network (RPN)
and second-stage losses of the Faster-RCNN. We adapt class-agnostic bounding box
regression model in the second-stage to get one bounding box per feature
proposal regardless of the number of base classes. Once trained, we
use the trained Faster-RCNN to extract $P$ bounding box proposals
$B\in\R^{P\times 4}$ and their corresponding d-dimensional features
$F\in\R^{P\times d}$ from each input image $\Ib$. We also apply an $\ell_2$ normalization layer to project all the features to the unit hypersphere. As discussed later, the normalization step is important as our model uses cosine similarity measure for better generalization.

We need the feature extracted from the full image bounding box to initialize our COL method. This is accomplished by manually appending the full image bounding box to the box proposals of the RPN, thus its feature is
extracted by the Faster-RCNN second-stage feature extractor. 
We denote the first
proposal in $B$ and $F$, the complete image bounding box and its feature,
respectively. \\

\subsection{Statistical Model Assumptions}\label{sec:stat_model}
Since the support set $\DD_{\rm train}$ provided to the learner is limited,
it is crucial to employ proper learning biases in the model to combat
overfitting. 
Inspired by the success of prototypical
networks~\cite{snell2017prototypical}, we design our model based on the
assumption that features of each object class form a single cluster in the
embedding space. 
We propose to use directional data based on the von Mises-Fisher (vMF) distribution, which arises naturally when each cluster is distributed on the unit hypersphere.
Formally, we assume features of each foreground class follow vMF distribution with mean direction $\btheta$ and positive concentration parameter $\kappa$
\begin{equation}
    p^+_{\btheta}(\xb) = \frac{1}{Z} \exp \left(\kappa \btheta^\top\xb  \right) \text{ s.t. } \norm{\btheta} = 1~,
\end{equation}
where $Z$ is the normalizing constant and input $\xb\in \R^d$ is a unit vector, i.e., $\norm{\xb}=1$ or equivalently $\xb \in \Sbb^{d-1}$. 
We assume the concentration hyperparameter is constant and the same for all novel classes. 
In \cref{sec:localization}, we propose an expectation maximization algorithm to estimate the mean direction of a novel class from the support set.

We could also use Gaussian distribution for our model whose effect was analogous to using Euclidean distance. 
We empirically show that vMF provides superior results to Gaussian distribution when using pre-trained features. 
Our results support related works in supervised few-shot learning~\cite{qi2018low,gidaris2018dynamic} where using cosine similarity outperforms euclidean distance measure. 
The underlying reason for this is well-studied by Wang \etal~\cite{wang2017normface}; Softmax loss used in the pre-training tends to create a `radial' feature distribution where direction specifies the semantic classes while magnitude decides the classification confidence.

Additionally, a background class distribution is learned to steer the learner toward objects and away from background proposals. 
Let  
\begin{equation}\label{eq:bg_dist}
p_{\bomega}^-(\xb) = \frac{1}{U} u^-_\bomega(\xb)~,
\end{equation}
represent the background class distribution where $U$ is a constant normalizer. 
As the base dataset provides a reach set of examples for learning the background model, the background distribution is learned from the base dataset and remains fixed when evaluating on WSOD examples sampled from the novel data. 

To learn the background distribution, we collect a set of background proposals with low intersection-over-union (IoU) score ($<0.3$) to the objects within the base dataset and use maximum likelihood estimation in~\cite{banerjee2005clustering} to find the parameters of vMF distribution for the background data.

\subsection{COL}\label{sec:localization}
\begin{figure}[t]
\centering
\includegraphics[width=0.98\linewidth]{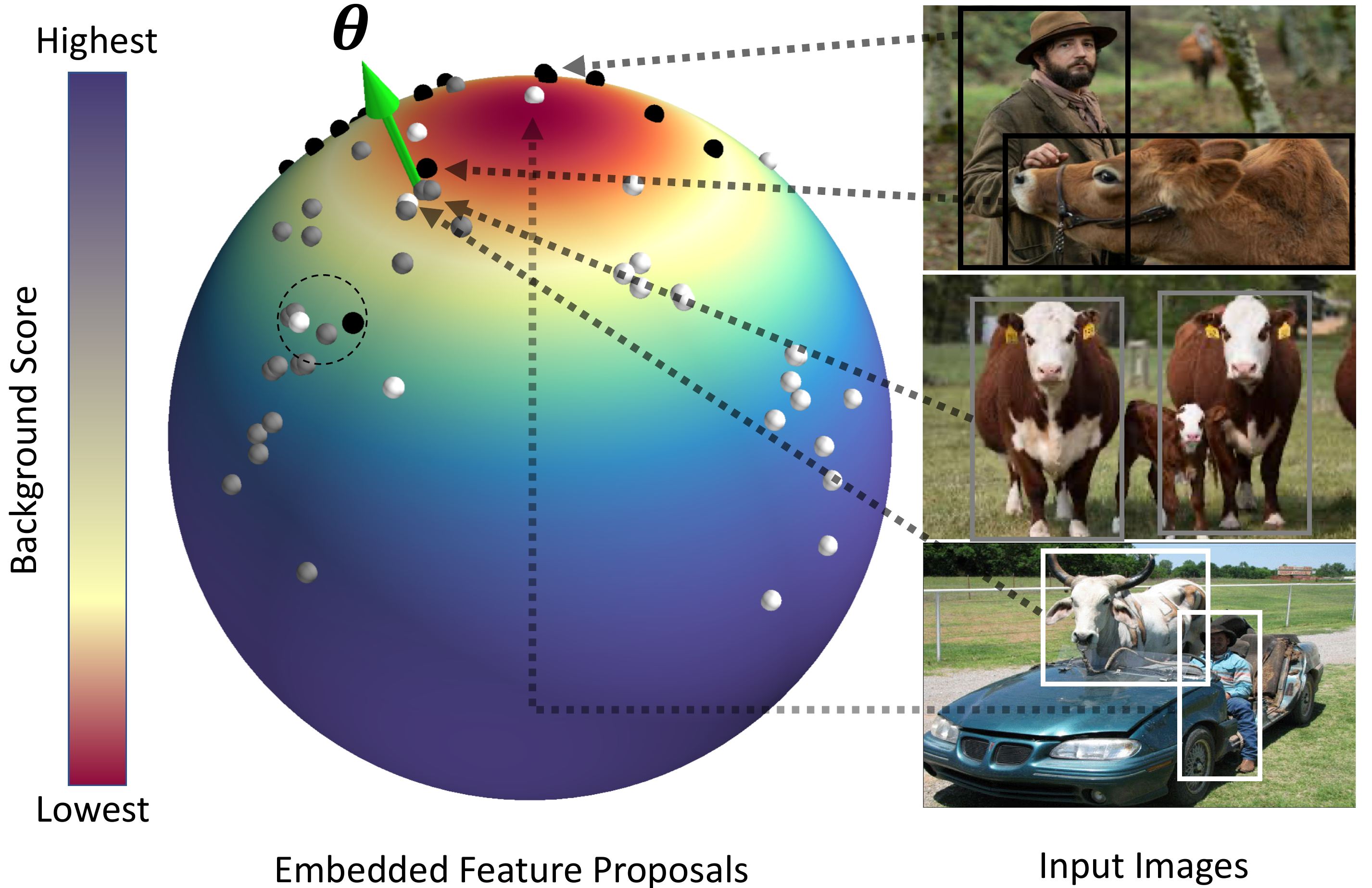}
\caption{\em\label{fig:em_goal} Example of COL across three images. 
Data points on the unit sphere represents feature proposals extracted from all input images. Features extracted from each image are colored the same (shown in white, gray, black colors). 
Background score function $u^-_{\bomega}(\xb)$ is also shown on unit sphere where blue and red indicated highest and lowest background scores, respectively.  
COL unit goal is to find a common object representation $\btheta$ (shown by green arrow) which is close to at least a white, gray, and black data point. Note that the area marked with dashed circle is also close to proposals from all three images but direction $\btheta$ is favored as it has a lower background score.}
\end{figure}

We first explain the method for few-shot COL with a single novel common object within 
the support set and employ it for few-shot WSOD later. 
As shown in \cref{fig:em_goal}, COL module's goal is to find the common object representation 
across a set of images with one novel object in common. Let $\FF = \{F_i\}_{i=1}^M$
denote the Faster-RCNN feature proposals extracted from the input images where $M$
is number of images. 
Each proposal has a (latent) binary label that indicates whether the proposal
tightly encloses the common object. 
Namely, $\zb_{ij} \in \{0, 1\}$ is the label
of the $j$-th proposal in the $i$-th image. 
Starting from an initial guess for the direction parameter $\btheta$ of the novel common class, the algorithm alternately refines the mean direction and label estimations in an expectation-maximization optimization framework. We present the update rules here and defer the derivations that bring interesting insights into the proposed method to \cref{sec:em_details}.
In the E-step, the algorithm uses the current direction to estimate soft labels $\wb$, where $\wb_{ik} \in [0, 1]$ is the soft label for the $k$-th proposal within the $i$-th image, via attention over the proposals within each image
\begin{equation}\label{eq:e_step}
\wb_{ik} = \frac{p^+_{\btheta}(F_{ik})/p_{\bomega}^-(F_{ik})}{\sum_{j=1}^P p^+_{\btheta}(F_{ij})/p_{\bomega}^-(F_{ij})}
\end{equation}
where $F_{ij}\in\Sbb^{d-1}$ is the feature of the $j$-th proposal in $F_i$. Recall that $p^+_{\btheta}$ and $p_{\bomega}^-$ are our foreground and learned background distributions introduced in \cref{sec:stat_model}. Note that it is unnecessary to know the normalization
factors $Z$ and $U$, since they cancel. In this step, proposal with a high foreground to background score ratio gets the highest label value within each image.

In the M-step, the mean direction $\btheta$ is updated given the new labels
\begin{equation}\label{eq:m_step}
\begin{split}
    \btheta &\gets \rm{norm}(\sum_{i=1}^M \tilde{\xb}_i)~, \\
    \text{where}\quad\tilde{\xb}_i &= \wb_i^\top F_i = \sum_{k=1}^P \wb_{ik} F_{ik}~, \\
\end{split}
\end{equation}
where $\rm{norm}(.)$ is $\ell_2$ normalization operator. 
Intuitively, one can see $\tilde{\xb}_i$ as the common object representation
within the $i$-th image which is estimated by computing the weighted average
over all the proposals where the contribution of proposals are controlled by
their soft labels. Given $\tilde{\xb}_i$, the novel class direction $\btheta$ is
estimated as the mean of the common object representations similar to the
prototypical networks~\cite{snell2017prototypical}. Finally, the estimated
mean is projected back onto the unit hypersphere.

\Cref{alg:coloc} summarizes our COL method. The problem is
solved in an iterative fashion by alternating between E-step in \cref{eq:e_step}
and M-step in \cref{eq:m_step} until convergence. 
Following the common practice
in WSOD~\cite{rahimi2020pairwise, uijlings2018revisiting,nguyen2009weakly}, we
use the bounding box feature extracted from the complete support images to initialize $\btheta$.
Recall that we use the first proposal in $F_i$ to represent the complete image
feature. Thus, the initialization step can be written as
\begin{equation}\label{eq:init}
    \btheta_{\rm init} \gets \rm{norm}(\sum_{i=1}^M F_{i1})~.
\end{equation}
\begin{algorithm}[t]
\DontPrintSemicolon
\KwIn{$\FF=\{F_1, \dots, F_M\}$, $u^-_\bomega$}
\KwOut{Common class mean direction $\btheta$}
$\btheta \gets \rm{norm}(\sum_{i=1}^M F_{i1})$\tcp*{Initialization}
\For(\tcp*[f]{Iterations}){$t \gets 1$ \textbf{to} T}{
    \For(\tcp*[f]{E-step}){$i \gets 1$ \textbf{to} M}{
        $\ob_{ij} \gets \kappa \btheta^\top F_{ij} - \log
u_{\bomega}^-(F_{ij})~~~\forall j\in [1, P]$\;
        $\wb_i \gets \text{softmax}(\ob_i)$\tcp*{Update Soft labels}
    }
    $\btheta \gets \rm{norm}(\sum_{i=1}^M \wb_i^\top F_i)$
\tcp*{M-step}
}
\caption{\label{alg:coloc} Common Object Localization}
\end{algorithm}
We remark that our initial direction is similar to what is used as class
mean in prototypical networks~\cite{snell2017prototypical}. What makes us
different is EM steps that refine the estimated mean by focusing on the common
objects and discarding background parts of the image.
\paragraph{Finding the Common Object in the Query Set} 
For a single feature proposal $\xb\in\Sbb^{d-1}$ extracted from query image
$\Ib$, our goal is to estimate the class label $c \in \{0,1\}$ which indicates if query proposal tightly encloses the target object. 
Given the estimated common object mean $\btheta$ and
the background class distribution $p_\bomega^-$, we compute
conditional class distribution function $P(c|\xb) \propto P(c) p(\xb|c)$ where
$P(c)$ and $p(\xb|c)$ are the class prior and likelihood, respectively. 
Let $P(c=1) = \alpha$, using the background and vMF
foreground class likelihoods, the conditional class distribution is written as
\begin{equation}
    P(c|\xb) \propto 
    \begin{cases}
    \exp\left(\kappa\btheta^\top\xb- \log u_\bomega^-(\xb)\right) &~~c=1 \\
    \lambda &~~ c = 0~,
    \end{cases}
\end{equation}
where $\lambda=(1-\alpha)/\alpha \times Z/U$ ($Z$ is vMF normalizer)
encapsulates all the constants. Equivalently, proposal $\xb$ can be classified
via a softmax over the logits 
\begin{equation}\label{eq:detection}
    {\rm logit}(c|\xb) =
    \begin{cases}
    \kappa\btheta^\top\xb  - \log u_\bomega^-(\xb) &~~c=1\\
    \log\lambda &~~ c = 0~.
    \end{cases}
\end{equation}
We set $\lambda = 1$ for all the COL experiments. Changing $\lambda$ adjusts the confidence values 
but keep the order of the final scores the same, therefore, its value does not affect the mean Average Precision (mAP) 
or Correct Localization (CorLoc) metrics. 

\subsection{WSOD}\label{sec:detection}
For the task of WSOD where we have more than one target class, our COL algorithm 
is first used to label instances of each class. Once the support set is labeled, an off-the-shelf 
few-shot object detection model can be used for learning novel classes. Inspired by the success of
the recent few-shot object detection method in~\cite{wang2020frustratingly}, we employ a single layer cosine similarity
classifier for learning.

Learning is performed on one target class $c \in \LL$ at a time. Let $\vb_c\in\R^d$ denote the classifier 
weight for class $c$. The classification score for this class is computed as
\begin{equation}\label{eq:wsod}
    s_c(\xb) = \frac{\tau \vb_c^\top\xb }{\norm{\vb_c}}~,
\end{equation}
where $\xb\in\Sbb^{d-1}$ is the $\ell_2$ normalized feature proposal extracted by our Faster-RCNN model 
and $\tau$ is temperature hyperparameter. 
For class $c$, the input training set $\DD_{\rm train}$ is split into 
positive images $\DD^c_{\rm train}$ of images that have the target class and 
negative set $\DD_{\rm train}\setminus\DD^c_{\rm train}$,
images without the target class. Then, we label $\DD^c_{\rm train}$ by running the COL algorithm on the positive
images and select the proposal with the highest soft label from each image as common object representative.
All the proposals in the negative set are used as negative examples. Finally, $\vb_c$ is learned
by minimizing the sigmoid cross entropy loss over the positive and negative proposals. We use L-BFGS
optimizer with strong Wolfe line search for faster convergence.

At the test time, test proposal
$\xb$ from the query set $\DD_{\rm test}$ is scored using the classifiers learned for each novel class.
\section{Related Work}
Multiple instance learning methods such as MI-SVM~\cite{andrews2003support} have been extensively used
for large-scale weakly supervised object detection. In a standard multiple instance learning framework,
latent bounding box labels and the appearance model are estimated jointly in an
alternating optimization process with the constraint that at least one bounding
box should be positive in each image. Alternating optimization combined with
modern deep neural network architecture is a dominant technique in the
literature showing the state-of-the-art performance in
WSOD~\cite{rahimi2020pairwise, uijlings2018revisiting,gokberk2014multi,siva2011weakly}. 
Ilse \etal~\cite{ilse2018attention} propose an attention-based
deep multiple instance learning architecture where bag label probability distribution is learned by
neural networks. More related to our work are a class of WSOD algorithms that
use knowledge transfer from a fully annotated base dataset to aid WSOD for a set
of novel classes~\cite{rahimi2020pairwise,uijlings2018revisiting,hoffman2016large,deselaers2010localizing}.
In~\cite{uijlings2018revisiting}, a class agnostic objectness score is learned
from the base dataset and is utilized to guide the multiple instance learning optimization steering
toward objects and away from the background. These methods rely on a relatively
large dataset to learn novel categories. 

Co-localization~\cite{li2016image}, co-segmentation~
\cite{li2018deep,vicente2011object}, and co-saliency~\cite{zhang2015co} methods
have the same kind of output as weakly-supervised object localization but they
typically do not utilize negative examples. More recently, several methods were
developed for localizing the common novel object under few-shot
setting~\cite{shaban2019learning,hu2019silco,siam2020weakly}. Shaban
\etal~\cite{shaban2019learning} learn a pairwise potential function between
proposals of the base classes and use this pairwise metric to solve a
minimum-energy labeling problem over a bidirectional graphical model to
co-localize novel classes. 
SILCO~\cite{hu2019silco} finds the common
object by computing a dense similarity map between each support image and the
query while only exploring the similarity among support images using their
coarse image-level features via a global average pooling. Although using global
average pooling reduces the computation, ignoring the dense similarities among
support images negatively affects the common object localization.

Few-shot learning has gained a lot of attention in image classification
~\cite{doersch2020crosstransformers,
rodriguez2020embedding,liu2019learning,finn2017model,vinyals2016matching}.
Prototypical networks~\cite{snell2017prototypical} use the mean of embedded
support examples to represent novel class prototypes and classifies query
examples by comparing their distances to the class prototypes. More recently,
Qi \etal~\cite{qi2018low} propose a weight imprinting process to learn the prototypes
on the unit hypersphere. Learning on the unit hypersphere has been employed by 
 other few-shot learning algorithms for better generalization~\cite{gidaris2018dynamic}
and stabilizing the training~\cite{wang2020frustratingly}. 
Most recently, Yang \etal~\cite{yang2021free} propose to make 
the distributions more Gaussian like
by transforming the features of the support set and query set using Tukey’s Ladder of 
Powers transformation~\cite{Tukey:107005}. It is shown that Tukey's normalization significantly improves 
the performance of few-shot prototypical learning. 
As the scope of these methods is limited to supervised learning,
we compare different normalization and transformations used in the literature 
for the weakly-supervised task of COL in \cref{sec:ablation}.

\section{Experiments}\label{sec:experiments}
We evaluate the proposed method in few-shot COL and WSOD
problems. 
We compare our work (vMF-MIL) with
Greedy Tree~\cite{shaban2019learning} and SILCO~\cite{hu2019silco}, two
state-of-the-art methods for the task of few-shot common object localization. 

To the best of our knowledge there is no WSOD algorithm for few-shot setting in
the literature. However, WSOD with knowledge transfer
methods~\cite{rahimi2020pairwise, uijlings2018revisiting,hoffman2016large,deselaers2010localizing} 
are closely related to our work. We
describe a slightly modified version of~\cite{uijlings2018revisiting}, called MI-SVM in our experiments, in \cref{sec:mi-svm} 
and discuss its differences to the proposed method. The MI-SVM baseline is not applicable to COL problem
as it always requires negative examples for training. To compare MI-SVM against
other COL methods, we provide MI-SVM with an extra set of
$K$ negative images that do not have the target class when sampling the support
set.

The original version of Greedy Tree selects only one proposal from each image in
the support set and does not perform detection on a new query image.
To make it compatible
with other methods, we add a simple inference step to the Greedy Tree. 
Let $\OO
= \{\xb_1, \dots, \xb_M\}$ denote the set of selected proposals, one from each
image in the support set. 
We score feature proposal $\xb$ from the query image
as a sum of its pairwise similarities to all the selected proposals, i.e.,
$\text{score}(\xb) = \sum_{j=1}^M r(\xb, \xb_j)$, where $r$ is the learned
pairwise similarity function by Greedy Tree. 
The computed score measures the
negative change in the energy value if $\xb$ were added as a new node to the
graph labeling problem used in~\cite{shaban2019learning}. 

In all the methods, we first hold out $20$ base classes for validation and
hyperparameter tuning and then re-train on all the base classes with the best
found parameters. 
For evaluation, we compute the correct localization (CorLoc)
rate~\cite{deselaers2010localizing} and mean Average Precision (mAP) with IoU
overlap threshold of $0.5$ on the query image.


%
%

\subsection{Common Object Localization}\label{sec:coloc}
\begin{table*}[!ht]
\centering
\caption{\em CorLoc ({\bf \em top}) and mAP ({\bf \em bottom}) performance of different few-shot common object localization methods on VOC07 test set. All of the models are trained on COCO60 and evaluated on a test query with $K=5$ images in the support set. The best and second best performing methods are shown in {\bf \em bold} and \colorbox[rgb]{0.92,0.92,0.92}{gray} background respectively. $^*$MI-SVM receives $K$ extra negative images.}
\label{tab:voc07_colcoK5_mAP}
\vspace{0.1cm}
\resizebox{\textwidth}{!}{%
\begin{tabular}{c||cccccccccccccccccccc|c}
\toprule
method      & aero & bike & bird & boat & bottle & bus  & car  & cat  & chair & cow  & table & dog  & horse & mbike & person & plant & sheep & sofa & train & tv & CorLoc  \\
\midrule
MI-SVM$^*$~\cite{uijlings2018revisiting} & $29.4$ & $13.2$ & {\cellcolor[rgb]{0.920,0.920,0.920}}$53.7$ & $32.7$ & {\cellcolor[rgb]{0.920,0.920,0.920}}$12.9$ & $70.4$ & $66.4$ & {\cellcolor[rgb]{0.920,0.920,0.920}}$67.4$ & $15.7$ & $81.6$ & $10.0$ & ${\bf 67.6}$ & $67.1$ & $27.1$ & $10.1$ & ${\bf 16.7}$ & $84.2$ & $38.9$ & $43.5$ & {\cellcolor[rgb]{0.920,0.920,0.920}}$41.1$ & $42.5$ \\
SILCO~\cite{hu2019silco}       & {\cellcolor[rgb]{0.920,0.920,0.920}}$51.0$      & {\cellcolor[rgb]{0.920,0.920,0.920}}$30.3$    & $50.7$ & {\cellcolor[rgb]{0.920,0.920,0.920}}$34.5$ & $11.3$   & {\cellcolor[rgb]{0.920,0.920,0.920}}$72.2$ & $63.6$ & $58.9$ & $11.2$  & $86.8$ & $6.7$         & $56.9$ & $51.9$  & ${\bf 49.2}$      & {\cellcolor[rgb]{0.920,0.920,0.920}}$13.0$   & ${\bf 16.7}$        & $52.6$  & {\cellcolor[rgb]{0.920,0.920,0.920}}$41.1$ & $46.8$  & $34.2$      & $42.0$   \\
Greedy Tree~\cite{shaban2019learning} & $35.3$      & $21.1$    & ${\bf 59.7}$ & {\cellcolor[rgb]{0.920,0.920,0.920}}$34.5$ & ${\bf 24.2}$   & ${\bf 77.8}$ & {\cellcolor[rgb]{0.920,0.920,0.920}}$73.4$ & $61.1$ & ${\bf 23.1}$  & {\cellcolor[rgb]{0.920,0.920,0.920}}$89.5$ & {\cellcolor[rgb]{0.920,0.920,0.920}}$15.0$        & $64.7$ & {\cellcolor[rgb]{0.920,0.920,0.920}}$73.4$  & $25.4$      & $12.8$   & $13.3$        & ${\bf 100.0}$ & ${\bf 64.2}$ & {\cellcolor[rgb]{0.920,0.920,0.920}}$61.3$  & ${\bf 46.6}$      & {\cellcolor[rgb]{0.920,0.920,0.920}}$48.8$   \\
vMF-MIL (ours) & ${\bf 62.7}$ & ${\bf 42.1}$ & {\cellcolor[rgb]{0.920,0.920,0.920}}$53.7$ & ${\bf 49.1}$ & $6.5$ & $68.5$ & ${\bf 73.8}$ & ${\bf 69.5}$ & {\cellcolor[rgb]{0.920,0.920,0.920}}$19.4$ & ${\bf 97.4}$ & ${\bf 36.7}$ & {\cellcolor[rgb]{0.920,0.920,0.920}}$65.7$ & ${\bf 82.3}$ & {\cellcolor[rgb]{0.920,0.920,0.920}}$40.7$ & ${\bf 21.7}$ & {\cellcolor[rgb]{0.920,0.920,0.920}}$15.0$ & {\cellcolor[rgb]{0.920,0.920,0.920}}$94.7$  & ${\bf 64.2}$ & ${\bf 69.4}$ & $31.5$ & ${\bf 53.2}$ \\
\midrule 
\midrule
method      & aero & bike & bird & boat & bottle & bus  & car  & cat  & chair & cow  & table & dog  & horse & mbike & person & plant & sheep & sofa & train & tv & mAP \\
\midrule
MI-SVM$^*$~\cite{uijlings2018revisiting} & $17.7$ & $7.7$  & $31.6$ & $10.6$ & {\cellcolor[rgb]{0.920,0.920,0.920}}$4.3$  & $46.5$ & $40.1$ & $53.3$ & $3.6$  & $56.8$ & $3.3$  & ${\bf 56.3}$ & $42.3$ & $17.1$ & $1.7$  & ${\bf 8.1}$  & $37.9$ & $25.9$ & $27.3$ & $19.2$ & $25.6$ \\
SILCO~\cite{hu2019silco}       & {\cellcolor[rgb]{0.920,0.920,0.920}}$33.0$      & {\cellcolor[rgb]{0.920,0.920,0.920}}$13.4$    & $34.5$ & ${\bf 14.8}$ & $3.9$    & $48.8$ & $38.4$ & {\cellcolor[rgb]{0.920,0.920,0.920}}$55.5$ & {\cellcolor[rgb]{0.920,0.920,0.920}}$4.0$   & $52.8$ & {\cellcolor[rgb]{0.920,0.920,0.920}}$4.5$         & $54.2$ & $36.3$  & ${\bf 27.3}$      & {\cellcolor[rgb]{0.920,0.920,0.920}}$3.3$    & {\cellcolor[rgb]{0.920,0.920,0.920}}$7.7$         & $27.0$  & $31.3$ & $36.7$  & {\cellcolor[rgb]{0.920,0.920,0.920}}$23.5$      & $27.5$ \\
Greedy Tree~\cite{shaban2019learning} & $26.0$      & $8.7$     & {\cellcolor[rgb]{0.920,0.920,0.920}}$37.4$ & $11.5$ & ${\bf 7.5}$    & {\cellcolor[rgb]{0.920,0.920,0.920}}$52.4$ & {\cellcolor[rgb]{0.920,0.920,0.920}}$47.7$ & $45.8$ & ${\bf 7.4}$   & {\cellcolor[rgb]{0.920,0.920,0.920}}$61.1$ & {\cellcolor[rgb]{0.920,0.920,0.920}}$4.5$         & $47.3$ & {\cellcolor[rgb]{0.920,0.920,0.920}}$50.7$  & $15.6$      & $2.3$    & $5.0$         & ${\bf 40.0}$  & ${\bf 46.3}$ & {\cellcolor[rgb]{0.920,0.920,0.920}}$47.3$  & ${\bf 25.7}$      & {\cellcolor[rgb]{0.920,0.920,0.920}}$29.5$ \\
vMF-MIL (ours) & ${\bf 36.7}$ & ${\bf 20.6}$ & ${\bf 38.1}$ & {\cellcolor[rgb]{0.920,0.920,0.920}}$14.2$ & $1.9$ & ${\bf 55.4}$ & ${\bf 50.2}$ & ${\bf 56.5}$ & ${\bf 7.4}$ & ${\bf 71.4}$ & ${\bf 9.5}$  & {\cellcolor[rgb]{0.920,0.920,0.920}}$56.2$ & ${\bf 63.4}$ & {\cellcolor[rgb]{0.920,0.920,0.920}}$16.8$ & ${\bf 4.9}$ & $3.4$ & {\cellcolor[rgb]{0.920,0.920,0.920}}$39.3$ & {\cellcolor[rgb]{0.920,0.920,0.920}}$43.0$ & ${\bf 51.2}$ & $23.0$ & ${\bf 33.1}$ \\
\bottomrule
\end{tabular}%
}
\end{table*}
\begin{table}[]
\caption{\em CorLoc(\%) and mAP(\%) results of different methods for the task of common object localization on novel object classes on the COCO60 dataset with support set size $K=5$ and $K=10$. $^*$MI-SVM receives $K$ extra negative images.}
\label{tab:coco60_coloc}
\vspace{0.1cm}
\resizebox{1.0\linewidth}{!}{%
\begin{tabular}{c||cc|cc}
\toprule
\multicolumn{1}{c}{\multirow{2}{*}{Model}} & \multicolumn{2}{c}{K = 5}   & \multicolumn{2}{c}{K=10} \\
\multicolumn{1}{c}{} & \multicolumn{1}{c}{CorLoc@0.5} & \multicolumn{1}{c}{mAP@0.5} & \multicolumn{1}{c}{CorLoc@0.5} & \multicolumn{1}{c}{mAP@0.5} \\
\midrule
MI-SVM$^*$~\cite{uijlings2018revisiting} & $30.5$ & $15.9$ & $33.2$ & $16.2$ \\
SILCO~\cite{hu2019silco} & $29.7$ & $14.8$ & $31.3$    & $15.8$    \\
Greedy Tree~\cite{shaban2019learning} & $32.7$ & $16.0$ & $33.8$ & $16.4$ \\
vMF-MIL (ours)  & ${\bf 34.8}$ & ${\bf 18.6}$ & ${\bf 36.9}$ & ${\bf 20.0}$ \\
\bottomrule 
\end{tabular}
}
\end{table}
We use the official implementations of SILCO and Greedy Tree for this
experiment. 
To have a fair comparison with SILCO, we employ Faster-RCNN with a
VGG16~\cite{simonyan2014very} backbone architecture for feature extraction in both Greedy Tree and our
method. 

We evaluate on a popular MS COCO 2014~\cite{lin2014microsoft} split used in
few-shot object detection methods~\cite{wang2020frustratingly,perez2020incremental, xiao2020few, yan2019meta, kang2019few}, named COCO60. 
In
COCO60 split, $60$ categories disjoint with the PASCAL VOC dataset are used as
base classes and the remaining $20$ classes are used as novel classes. 
This
allows us to also perform a cross-dataset evaluation on PASCAL
VOC07~\cite{pascal-voc-2007} test set. 
We evaluate the performance of each
method over $2000$ randomly sampled tasks.

\Cref{tab:voc07_colcoK5_mAP} and \cref{tab:coco60_coloc} summarize the results
on PASCAL VOC and MS COCO datasets, respectively. 
Despite its simplicity, our
method outperforms all the methods by a large margin, followed by Greedy Tree
and SILCO. 
Specifically, we gain between $10\%$ to $20\%$ relative improvement in mAP metric against the second best performing method. 
The proposed method and Greedy Tree both
estimate latent proposal-level labels of the support images to find the common
object. 
However, SILCO explores the dense similarity between each support image
and the query image while using coarse image-level features via a global average
pooling to estimate the relation of support images. 
This experiment confirms
estimating proposal-level labels within the support images is quite important
for common object localization.


\subsubsection{Direct Comparison to Greedy Tree}
To ensure a fair comparison, we also compare our common object localization unit
to the Greedy Tree by exactly following the original experimental protocol
in~\cite{shaban2019learning}. 
Their algorithm utilizes a split of the COCO 2017
dataset with $63$ base classes for training and $17$ held-out novel classes for
testing the algorithm. 
The trained model is also tested on a subset of the ILSVRC
2013 detection dataset with $148$ novel classes that have no overlap with the base
classes. 
In Greedy Tree, a Faster-RCNN with ResNet50~\cite{he2016deep} backbone
is first trained on the base classes and used to extract features from all the
images. 
To allow a fair comparison, we use the same feature set provided by the
authors. 
To mimic common object localization during training, we sample
tasks with $N=1$ and $K=8$ for training.

Similar to~\cite{shaban2019learning}, we evaluate our model over $1000$ randomly
sampled tasks each containing $K=8$ images with an object class in common. 
For
each image, the proposal with the highest soft label in \cref{eq:e_step} is
returned as the common object. 
We report the class-agnostic CorLoc ratio on COCO
and ILSVRC datasets in \cref{tab:coco63_ilsvrc13_agcorloc} and compare it with
the results in~\cite{shaban2019learning}. 
vMF-MIL outperforms Greedy Tree by
$2.20\%$ and $1.75\%$ in MS COCO and ILSVRC datasets, respectively.

\begin{table}[h]
\centering
\caption{\em Class-agnostic CorLoc(\%) with $95\%$ confidence interval of the methods in~\cite{shaban2019learning} compared to our method. All methods use $K=8$ positive images for finding the common object.}
\label{tab:coco63_ilsvrc13_agcorloc}
\vspace{0.1cm}
\resizebox{0.8\linewidth}{!}{%
\begin{tabular}{c||c|c}
\toprule
method      & COCO           & ILSVRC13         \\
\midrule
TRWS~\cite{shaban2019learning}        & $64.53 \pm 1.05$ & $52.95 \pm 1.09$ \\
ASTAR~\cite{shaban2019learning}       & $64.54 \pm 1.05$ & $52.89 \pm 1.09$ \\
Greedy Tree~\cite{shaban2019learning} & $64.65 \pm 1.05$ & $53.00 \pm 1.10$ \\
vMF-MIL (ours)        & ${\bf 66.85 \pm 1.03}$ & ${\bf 54.75 \pm 1.09}$ \\
\bottomrule
\end{tabular}%
}
\vspace{-0.2cm}
\end{table}

\subsection{Few-shot WSOD}\label{sec:fswsod}
We train our model on COCO60 for the task of few-shot WSOD with different
$N$-way, $K$-shot problems and compare it with the knowledge transfer MI-SVM model
described in \cref{sec:mi-svm} on PASCAL VOC 2007 and MS COCO novel classes in
\cref{tab:fswsod}. 
To highlight the importance of EM refinement, we also train
our model with full image prototypical initialization without EM
refinement. 
In both datasets, vMF-MIL outperforms MI-SVM in all the scenarios,
demonstrating the strong generalization ability of our learning approach.  


\begin{table}[h]
\centering
\caption{\em mAP(\%) of different few-shot WSOD methods on COCO60 and PASCAL VOC datasets.}
\label{tab:fswsod}
\vspace{0.1cm}
\resizebox{1.0\linewidth}{!}{%
\begin{tabular}{c|c||cc|cc|cc}
\toprule
\multirow{2}{*}{Method} & \multirow{2}{*}{Dataset} & \multicolumn{2}{c}{$N = 5$} & \multicolumn{2}{c}{$N=10$} & \multicolumn{2}{c}{$N=20$} \\
               &                         & $K=5$                       & K=10    & K=5    & K=10   & K=5    & K=10   \\
\midrule
Prototypical Init  & \multirow{3}{*}{VOC07} & $16.01$                         & $17.93$     & $10.56$    & $11.02$    & $5.41$    & $6.72$    \\
MI-SVM~\cite{uijlings2018revisiting}         &                         & $17.99$                         & $20.27$     & $12.09$    & $13.07$    & $7.04$    & $8.32$    \\
vMF-MIL (ours) &                         & ${\bf 21.22}$                         & ${\bf 22.01}$     & ${\bf 14.54}$    & ${\bf 15.83}$    & ${\bf 8.83}$    & ${\bf 10.19}$   \\
\midrule 
\midrule
Prototypical Init  & \multirow{3}{*}{COCO60} & $8.90$                         & $9.28$     & $4.65$    & $6.07$    & $2.99$    & $3.26$    \\
MI-SVM~\cite{uijlings2018revisiting}         &                         & \multicolumn{1}{l}{$11.40$} & $11.60$ & $7.30$ & $7.80$ & $2.97$ & $3.70$ \\
vMF-MIL (ours) &                         & \multicolumn{1}{l}{${\bf 12.35}$} & ${\bf 13.19}$ & ${\bf 8.53}$ & ${\bf 10.07}$ & ${\bf 4.23}$ & ${\bf 4.85}$ \\
\bottomrule
\end{tabular}%
}
\end{table}
\begin{figure}[h]
\begin{center}
\includegraphics[width=0.8\linewidth]{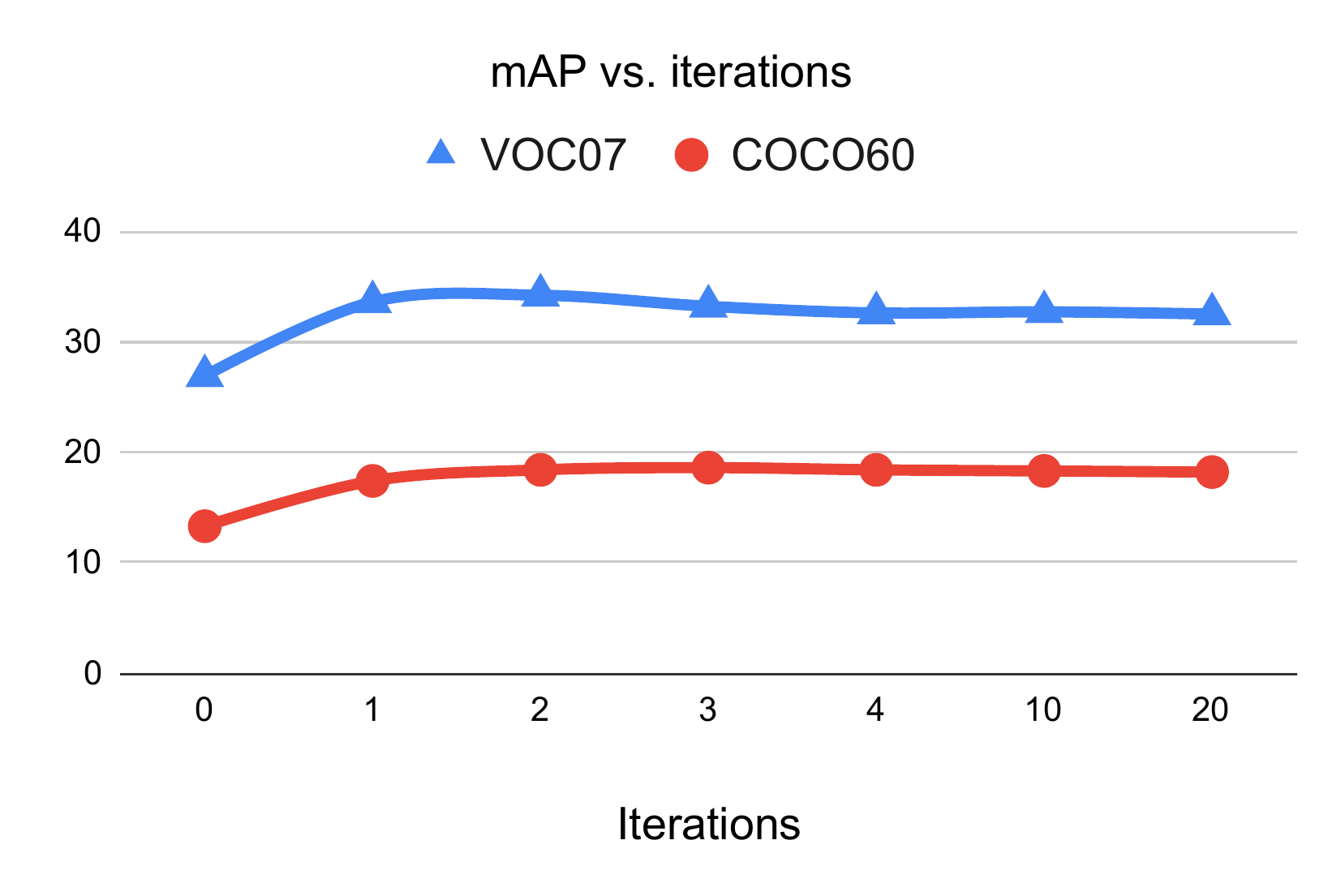}
\caption{\em \label{fig:iters}  mAP(\%) vs. number of EM iterations in
common object localization task with $K=5$ on COCO60 and VOC07 datasets. The performance
reaches a plateau at step $4$.}
\vspace{-0.8cm}
\end{center}
\end{figure}

\subsection{Large-Scale WSOD}\label{sec:largescale}
Although our method is designed for low-shot settings, it is interesting to evaluate its performance in the standard WSOD setting as well.
Related to our work is transfer learning approaches for large-scale WSOD~\cite{uijlings2018revisiting,hoffman2016large} with ImageNet detection as the standard benchmark.
Typically, the first $100$ classes are used as the base dataset and the rest $100$ with $65$k images are used as novel objects. 
We follow the setup in~\cite{uijlings2018revisiting} and use the pre-trained Inception-Resnet Faster-RCNN model and weights provided by the authors to extract proposals, features, and the objectness scores. 
We apply our EM algorithm to the extracted features and use $u^-_\bomega(\xb)=\alpha(1-\rm{obj}(\xb))$ for each proposal $\xb$ where $\rm{obj}(\xb)$ is the Faster-RCNN objectness score and $\alpha$ is a hyper-parameter we tuned for the task. 
To our surprise, vFM-MIL outperforms~\cite{uijlings2018revisiting} in Table~\ref{tab:largescale} while being about $100\times$ faster. We believe this results can be further improved by relaxing some of the assumptions in our statistical model as overfitting may not be as significant in large-scale. For instance, we can learn a separate concentration parameter for each novel class in the EM steps. Furthermore, we can utilize the novel dataset to update the background scoring function $u^-_\omega$. We defer these improvements and further analysis to the future work.
\begin{table}[h]
\centering
\caption{\label{tab:largescale}\em\small Large-Scale WSOD on ImageNet Detection.}
\resizebox{1.0\linewidth}{!}{%
\begin{tabular}{c||c|c}
\toprule
Model & CorLoc@0.5 & Time (min.)  \\
\midrule
LSDA (JMLR 2016)~~\cite{hoffman2016large} & $28.8$ & - \\
Uijlings~et~al. (CVPR~18)~~\cite{uijlings2018revisiting}& $74.2$ & $900$ (estimated) \\
vMF-MIL (ours) & $\mathbf{76.5}$ & $10$ \\
\bottomrule
\end{tabular}%
}
\end{table}
\subsection{Ablation Study}\label{sec:ablation}
To understand which parts of the proposed method are critical for common
object localization, we analyzed results in \cref{tab:coco60_coloc} with $K=5$
for each of the important components of the proposed method in
\cref{tab:variants}. 
These components are: initializing $\btheta$ with the mean
of complete image features (Prototypical Init), EM
updates, and learning background distribution $p_\bomega^-$ to steer the
algorithm toward objects.
The first entry (\#1) in \cref{tab:variants} shows that there is a huge performance gap when the background model is not used.
This is expected,
since without using the background model it may localize non-object patterns
such as grass, water, building, etc. with similar appearances as the common object.
Comparing the third entry (\#3) with the complete model (\#5) reveals that using EM
refinements alone increases CorLoc by $3.9\%$ and mAP by $5.3\%$. 
Finally, the
fourth entry shows that the EM steps are only effective if $\btheta$ is initialized with the complete image proposal. 

Second part of \cref{tab:variants} shows the advantage of using vMF to Gaussian distribution in the EM algorithm (see \cref{sec:gaussian} for the details). Tukey's transformation~\cite{Tukey:107005} furthur improves the performance of Gaussian model but vMF distribution still exhibits the best performance. 
We believe this is because feature vectors' direction better capture the semantic information.
\begin{table}[h]
\centering
\caption{\em Ablation study on COCO60 dataset. \#1-5 show the importance of initialization, iterative EM updates, and learning the background model. \#6-8 compare different statistical models in the EM algorithm.}
\label{tab:variants}
\vspace{0.1cm}
\resizebox{1.0\linewidth}{!}{%
\begin{tabular}{c|cccc|cc}
\toprule
\# & Random Init & Prototypical Init & EM Updates & $p_\bomega^-$ & CorLoc & mAP  \\
\midrule
1&&\checkmark & \checkmark &             & $1.9$  & $0.6$  \\
2&&           &            & \checkmark  & $22.8$ & $9.3$ \\
3&&\checkmark &            & \checkmark  & $30.9$ & $13.3$ \\
4&\checkmark & & \checkmark & \checkmark  & $30.1$ & $14.2$ \\
5&&\checkmark & \checkmark & \checkmark  & ${\bf 34.8}$ & ${\bf 18.6}$ \\
\midrule\midrule
 & Gaussian & Tukey+Gaussian & vMF & &  &   \\
\midrule
6&\checkmark&&&& $29.8$ & $13.9$ \\
7&&\checkmark&&& $34.0$ & $17.3$ \\
8&&&\checkmark&& ${\bf 34.8}$ & ${\bf 18.6}$ \\
\bottomrule
\end{tabular}%
}
\end{table}

\begin{figure}[t]
\begin{center}
\includegraphics[width=0.9\linewidth]{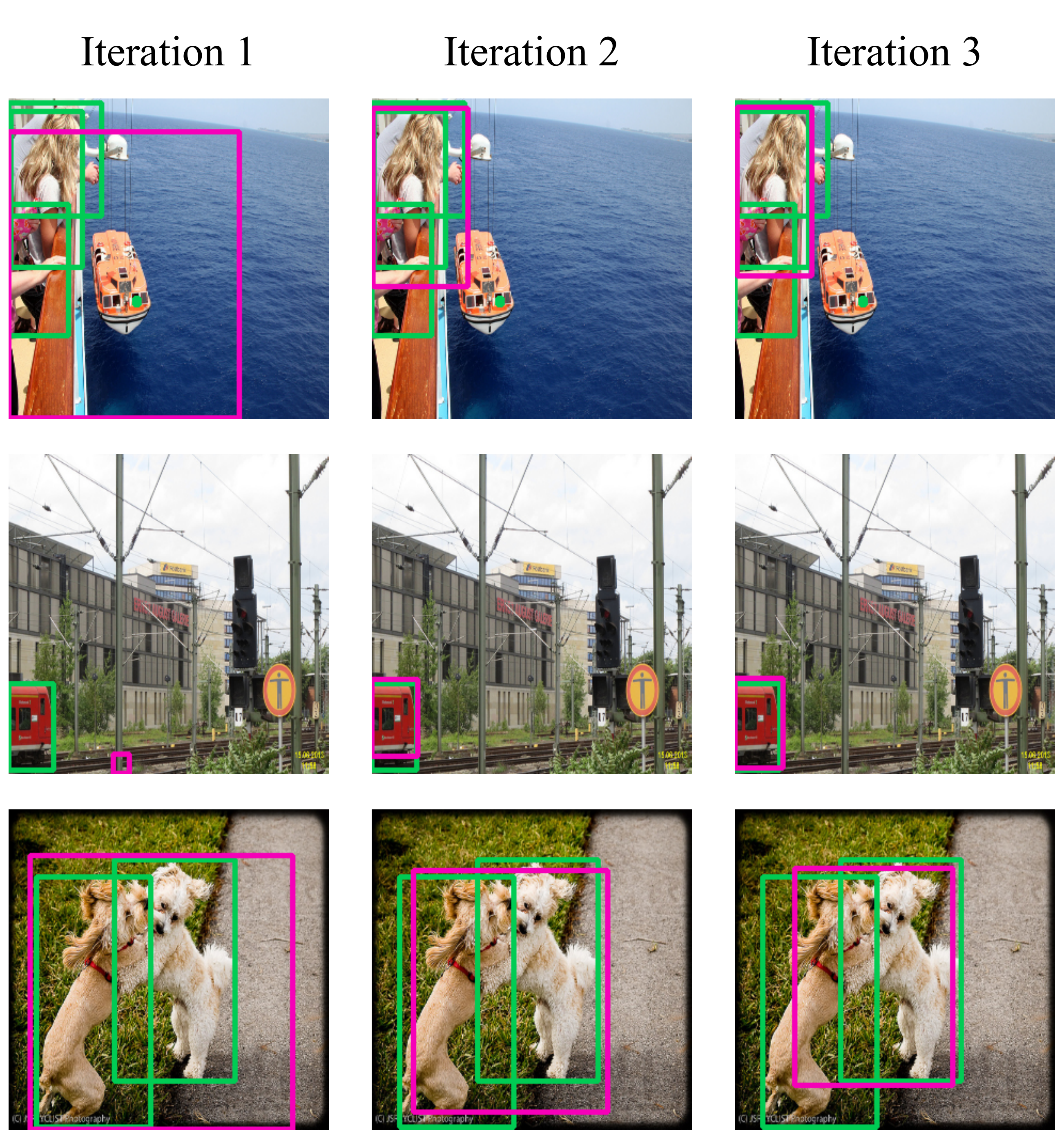}
\caption{\em \label{fig:images_iters} Bounding box adjustments at each iteration for
the common object localization experiment on COCO60 with $K=5$. Only the top
prediction in the query image is shown(in pink color) for each iteration. Ground-truth bounding boxes of the target classes are shown in green. EM refinements improve the target object localization in query image.}
\end{center}
\vspace{-0.8cm}
\end{figure}

Finally, we illustrate the performance improvement vs. the number of EM
steps in \cref{fig:iters}. 
In both VOC07 and COCO60 datasets, mAP reaches a
plateau showing that the algorithm converges quickly. 
Qualitative results in
\cref{fig:images_iters} depict successful cases where EM refinements improve the
top prediction.

\section{Conclusion}
We have presented vMF-MIL, a multiple instance learning framework to address
the problem of few-shot common object localization and WSOD. vMF-MIL uses a
simple inductive bias in learning to combat the overfitting issue in few-shot
learning.
Specifically, instances of each class are assumed to form a cluster on a unit hypersphere, 
where the mean corresponds to the class prototype. 
Our experiments on few-shot common object localization illustrate the advantage
of our simple approach over several state-of-the-art methods, improving
the few-shot WSOD performance compared with the strong MI-SVM baseline.

\clearpage
{\small
\bibliographystyle{ieee_fullname}
\bibliography{egbib}

\begin{thebibliography}{10}\itemsep=-1pt

\bibitem{amos2017optnet}
Brandon Amos and J.~Zico Kolter.
\newblock {O}pt{N}et: Differentiable optimization as a layer in neural
  networks.
\newblock In {\em Int. Conf. Mach. Learn.}, 2017.

\bibitem{andrews2003support}
Stuart Andrews, Ioannis Tsochantaridis, and Thomas Hofmann.
\newblock Support vector machines for multiple-instance learning.
\newblock In {\em Adv. Neural Inform. Process. Syst.}, 2003.

\bibitem{banerjee2005clustering}
Arindam Banerjee, Inderjit~S Dhillon, Joydeep Ghosh, Suvrit Sra, and Greg
  Ridgeway.
\newblock Clustering on the unit hypersphere using von mises-fisher
  distributions.
\newblock {\em Journal of Machine Learning Research}, 2005.

\bibitem{imagenet_cvpr09}
J. Deng, W. Dong, R. Socher, L.-J. Li, K. Li, and L. Fei-Fei.
\newblock {ImageNet: A Large-Scale Hierarchical Image Database}.
\newblock In {\em IEEE Conf. Comput. Vis. Pattern Recog.}, 2009.

\bibitem{deselaers2010localizing}
Thomas Deselaers, Bogdan Alexe, and Vittorio Ferrari.
\newblock Localizing objects while learning their appearance.
\newblock In {\em Eur. Conf. Comput. Vis.}, 2010.

\bibitem{doersch2020crosstransformers}
Carl Doersch, Ankush Gupta, and Andrew Zisserman.
\newblock Crosstransformers: spatially-aware few-shot transfer.
\newblock {\em Adv. Neural Inform. Process. Syst.}, 2020.

\bibitem{pascal-voc-2007}
M. Everingham, L. Van~Gool, C.~K.~I. Williams, J. Winn, and A. Zisserman.
\newblock The {PASCAL} {V}isual {O}bject {C}lasses {C}hallenge 2007 {(VOC2007)}
  {R}esults.
\newblock
  http://www.pascal-network.org/challenges/VOC/voc2007/workshop/index.html.

\bibitem{finn2017model}
Chelsea Finn, Pieter Abbeel, and Sergey Levine.
\newblock Model-agnostic meta-learning for fast adaptation of deep networks.
\newblock {\em Int. Conf. Mach. Learn.}, 2017.

\bibitem{gidaris2018dynamic}
Spyros Gidaris and Nikos Komodakis.
\newblock Dynamic few-shot visual learning without forgetting.
\newblock In {\em IEEE Conf. Comput. Vis. Pattern Recog.}, 2018.

\bibitem{gokberk2014multi}
Ramazan Gokberk~Cinbis, Jakob Verbeek, and Cordelia Schmid.
\newblock Multi-fold mil training for weakly supervised object localization.
\newblock In {\em IEEE Conf. Comput. Vis. Pattern Recog.}, 2014.

\bibitem{he2016deep}
Kaiming He, Xiangyu Zhang, Shaoqing Ren, and Jian Sun.
\newblock Deep residual learning for image recognition.
\newblock In {\em IEEE Conf. Comput. Vis. Pattern Recog.}, 2016.

\bibitem{hoffman2016large}
Judy Hoffman, Deepak Pathak, Eric Tzeng, Jonathan Long, Sergio Guadarrama,
  Trevor Darrell, and Kate Saenko.
\newblock Large scale visual recognition through adaptation using joint
  representation and multiple instance learning.
\newblock {\em The Journal of Machine Learning Research}, 2016.

\bibitem{hu2019silco}
Tao Hu, Pascal Mettes, Jia-Hong Huang, and Cees~GM Snoek.
\newblock Silco: Show a few images, localize the common object.
\newblock In {\em Int. Conf. Comput. Vis.}, 2019.

\bibitem{ilse2018attention}
Maximilian Ilse, Jakub~M Tomczak, and Max Welling.
\newblock Attention-based deep multiple instance learning.
\newblock {\em Int. Conf. Mach. Learn.}, 2018.

\bibitem{kang2019few}
Bingyi Kang, Zhuang Liu, Xin Wang, Fisher Yu, Jiashi Feng, and T~revor Darrell.
\newblock Few-shot object detection via feature reweighting.
\newblock In {\em Int. Conf. Comput. Vis.}, 2019.

\bibitem{kim2020tell}
Daesik Kim, Gyujeong Lee, Jisoo Jeong, and Nojun Kwak.
\newblock Tell me what they're holding: Weakly-supervised object detection with
  transferable knowledge from human-object interaction.
\newblock In {\em AAAI}, 2020.

\bibitem{lee2019meta}
Kwonjoon Lee, Subhransu Maji, Avinash Ravichandran, and Stefano Soatto.
\newblock Meta-learning with differentiable convex optimization.
\newblock In {\em IEEE Conf. Comput. Vis. Pattern Recog.}, 2019.

\bibitem{li2018deep}
Weihao Li, Omid~Hosseini Jafari, and Carsten Rother.
\newblock Deep object co-segmentation.
\newblock In {\em ACCV}, 2018.

\bibitem{li2016image}
Yao Li, Lingqiao Liu, Chunhua Shen, and Anton van~den Hengel.
\newblock Image co-localization by mimicking a good detector’s confidence
  score distribution.
\newblock In {\em Eur. Conf. Comput. Vis.}, 2016.

\bibitem{lin2014microsoft}
Tsung-Yi Lin, Michael Maire, Serge Belongie, James Hays, Pietro Perona, Deva
  Ramanan, Piotr Doll{\'a}r, and C~Lawrence Zitnick.
\newblock Microsoft coco: Common objects in context.
\newblock In {\em Eur. Conf. Comput. Vis.}, 2014.

\bibitem{liu2019learning}
Yanbin Liu, Juho Lee, Minseop Park, Saehoon Kim, Eunho Yang, Sung~Ju Hwang, and
  Yi Yang.
\newblock Learning to propagate labels: Transductive propagation network for
  few-shot learning.
\newblock {\em Int. Conf. Learn. Represent.}, 2019.

\bibitem{lu2019see}
Xiankai Lu, Wenguan Wang, Chao Ma, Jianbing Shen, Ling Shao, and Fatih Porikli.
\newblock See more, know more: Unsupervised video object segmentation with
  co-attention siamese networks.
\newblock In {\em IEEE Conf. Comput. Vis. Pattern Recog.}, 2019.

\bibitem{mardia2009directional}
Kanti~V Mardia and Peter~E Jupp.
\newblock {\em Directional statistics}.
\newblock John Wiley \& Sons, 2009.

\bibitem{maron1998framework}
Oded Maron and Tom{\'a}s Lozano-P{\'e}rez.
\newblock A framework for multiple-instance learning.
\newblock In {\em Adv. Neural Inform. Process. Syst.}, 1998.

\bibitem{nguyen2009weakly}
Minh~Hoai Nguyen, Lorenzo Torresani, Fernando De~La~Torre, and Carsten Rother.
\newblock Weakly supervised discriminative localization and classification: a
  joint learning process.
\newblock In {\em Int. Conf. Comput. Vis.}, 2009.

\bibitem{perez2020incremental}
Juan-Manuel Perez-Rua, Xiatian Zhu, Timothy~M Hospedales, and Tao Xiang.
\newblock Incremental few-shot object detection.
\newblock In {\em IEEE Conf. Comput. Vis. Pattern Recog.}, 2020.

\bibitem{qi2018low}
Hang Qi, Matthew Brown, and David~G Lowe.
\newblock Low-shot learning with imprinted weights.
\newblock In {\em IEEE Conf. Comput. Vis. Pattern Recog.}, 2018.

\bibitem{rahimi2020pairwise}
Amir Rahimi, Amirreza Shaban, Thalaiyasingam Ajanthan, Richard Hartley, and
  Byron Boots.
\newblock Pairwise similarity knowledge transfer for weakly supervised object
  localization.
\newblock In {\em Eur. Conf. Comput. Vis.}, 2020.

\bibitem{ren2016faster}
Shaoqing Ren, Kaiming He, Ross Girshick, and Jian Sun.
\newblock Faster r-cnn: Towards real-time object detection with region proposal
  networks.
\newblock {\em IEEE Trans. Pattern Anal. Mach. Intell.}, 2016.

\bibitem{rodriguez2020embedding}
Pau Rodr{\'\i}guez, Issam Laradji, Alexandre Drouin, and Alexandre Lacoste.
\newblock Embedding propagation: Smoother manifold for few-shot classification.
\newblock {\em Eur. Conf. Comput. Vis.}, 2020.

\bibitem{shaban2019learning}
Amirreza Shaban, Amir Rahimi, Shray Bansal, Stephen Gould, Byron Boots, and
  Richard Hartley.
\newblock Learning to find common objects across few image collections.
\newblock In {\em Int. Conf. Comput. Vis.}, 2019.

\bibitem{siam2020weakly}
Mennatullah Siam, Naren Doraiswamy, Boris~N Oreshkin, Hengshuai Yao, and Martin
  Jagersand.
\newblock Weakly supervised few-shot object segmentation using co-attention
  with visual and semantic embeddings.
\newblock {\em IJCAI}, 2020.

\bibitem{simonyan2014very}
Karen Simonyan and Andrew Zisserman.
\newblock Very deep convolutional networks for large-scale image recognition.
\newblock {\em Int. Conf. Learn. Represent.}, 2015.

\bibitem{siva2011weakly}
Parthipan Siva and Tao Xiang.
\newblock Weakly supervised object detector learning with model drift
  detection.
\newblock In {\em Int. Conf. Comput. Vis.}, 2011.

\bibitem{snell2017prototypical}
Jake Snell, Kevin Swersky, and Richard Zemel.
\newblock Prototypical networks for few-shot learning.
\newblock In {\em Adv. Neural Inform. Process. Syst.}, 2017.

\bibitem{Tukey:107005}
John~W Tukey.
\newblock {\em {Exploratory data analysis}}.
\newblock Addison-Wesley Series in Behavioral Science. Addison-Wesley, Reading,
  MA, 1977.

\bibitem{uijlings2018revisiting}
Jasper Uijlings, Stefan Popov, and Vittorio Ferrari.
\newblock Revisiting knowledge transfer for training object class detectors.
\newblock In {\em IEEE Conf. Comput. Vis. Pattern Recog.}, 2018.

\bibitem{vicente2011object}
Sara Vicente, Carsten Rother, and Vladimir Kolmogorov.
\newblock Object cosegmentation.
\newblock In {\em IEEE Conf. Comput. Vis. Pattern Recog.}, 2011.

\bibitem{vinyals2016matching}
Oriol Vinyals, Charles Blundell, Timothy Lillicrap, Daan Wierstra, et~al.
\newblock Matching networks for one shot learning.
\newblock In {\em Adv. Neural Inform. Process. Syst.}, 2016.

\bibitem{wang2017normface}
Feng Wang, Xiang Xiang, Jian Cheng, and Alan~Loddon Yuille.
\newblock Normface: L2 hypersphere embedding for face verification.
\newblock In {\em Proceedings of the 25th ACM international conference on
  Multimedia}, pages 1041--1049, 2017.

\bibitem{wang2020frustratingly}
Xin Wang, Thomas~E Huang, Trevor Darrell, Joseph~E Gonzalez, and Fisher Yu.
\newblock Frustratingly simple few-shot object detection.
\newblock {\em Int. Conf. Mach. Learn.}, 2020.

\bibitem{wang2019meta}
Yu-Xiong Wang, Deva Ramanan, and Martial Hebert.
\newblock Meta-learning to detect rare objects.
\newblock In {\em Int. Conf. Comput. Vis.}, 2019.

\bibitem{xiao2020few}
Yang Xiao and Renaud Marlet.
\newblock Few-shot object detection and viewpoint estimation for objects in the
  wild.
\newblock {\em Eur. Conf. Comput. Vis.}, 2020.

\bibitem{yan2019meta}
Xiaopeng Yan, Ziliang Chen, Anni Xu, Xiaoxi Wang, Xiaodan Liang, and Liang Lin.
\newblock Meta r-cnn: Towards general solver for instance-level low-shot
  learning.
\newblock In {\em Int. Conf. Comput. Vis.}, 2019.

\bibitem{yang2021free}
Shuo Yang, Lu Liu, and Min Xu.
\newblock Free lunch for few-shot learning: Distribution calibration.
\newblock {\em Int. Conf. Learn. Represent.}, 2021.

\bibitem{zhang2015co}
Dingwen Zhang, Junwei Han, Chao Li, and Jingdong Wang.
\newblock Co-saliency detection via looking deep and wide.
\newblock In {\em IEEE Conf. Comput. Vis. Pattern Recog.}, 2015.

\end{thebibliography}
}

\clearpage
\appendix
\onecolumn
\begin{center}
{\bf \Large Appendix} \\ \vspace{.3cm}
{\bf \large Few-shot Weakly-Supervised Object Detection via Directional Statistics}\\\vspace{.2cm}
\end{center}
\setcounter{page}{1}

\section{Expectation-Maximization Derivation}\label{sec:em_details}
Recall that each proposal has a (latent) binary label $\zb_{ij} \in \{0, 1\}$ that indicates whether the proposal
tightly encloses the common object. Following the best practice of the
previous works in WSOD~\cite{rahimi2020pairwise,
uijlings2018revisiting,gokberk2014multi}, we assume there is exactly one
proposal with label $1$ (positive proposal) in each image and the rest are
negative proposals, i.e., $\zb_i \in \{\eb_1, \dots \eb_P\}$ where
$\eb_j\in\{0,1\}^P$ is the $j$-th canonical basis. 

Assuming that the images are sampled
independently given the common class $c$, the full likelihood function is given by
\begin{equation}
\label{eq:full_likelihood}
\begin{split}
    l(\btheta; \FF) &= p(\FF|\btheta) = \prod_{i=1}^M p(F_i|\btheta)\\
    &= \prod_{i=1}^M \sum_{j=1}^P p(F_i, \zb_i=\eb_j|\btheta)~,
\end{split}
\end{equation}
where $\btheta\in \Sbb^{d-1}$ is the mean direction of the common class distribution.
Note that the last equation integrates over all possible values of $\zb_i$. 
Assuming that proposals are i.i.d samples from their
corresponding distributions given their labels $\zb_i$\footnote{i.i.d assumption is a standard approach in MIL and works well in practice. See \cite{maron1998framework} for more details.}, we can write
\begin{equation}
\label{eq:bag_dist_1}
    p(F_i|\zb_i=\eb_j, \btheta) = p_{\btheta}^+(F_{ij})
\prod_{\substack{k=1\\k\neq j}}^P p_{\bomega}^-(F_{ik})~,
\end{equation}
where $F_{ij}$ is the feature of the $j$-th proposal in $F_i$,
$p_{\btheta}^+$ is the generic distribution that generates the common class
proposals, and $p_{\bomega}^-$ represents background proposals' distribution.
For brevity, let us re-write \cref{eq:bag_dist_1} in a more compact form
\begin{equation}
\label{eq:bag_dist}
    p(F_i|\zb_i=\eb_j, \btheta) = q_{\btheta}(F_{ij}) p^-_{\bomega}(F_i)~,
\end{equation}
where $q_{\btheta}$ is quotient of object and background distributions
$q_{\btheta}(\xb) = p_{\btheta}^+(\xb)/p_{\bomega}^-(\xb)$, and
$p^-_{\bomega}(F_i) = \prod_{j=1}^P p_{\bomega}^-(F_{ij})$.

We adopt the EM algorithm to maximize the likelihood in
\cref{eq:full_likelihood} by iteratively optimizing the surrogate expected
log-likelihood which is easier to compute. In the E-step, the posterior
distribution of the latent variables $\wb_{ik} = p(\zb_i=\eb_k| F_i, \btheta)$
are computed for the current $\btheta$. By \cref{eq:bag_dist}, using Bayes'
theorem, and assuming a uniform distribution over image labels $\zb_i$, the
posterior can be expressed in terms of quotient of distributions defined above 
\begin{equation}\label{eq:e_step_app}
\wb_{ik} = \frac{q_{\btheta}(F_{ik})}{\sum_{j=1}^P q_{\btheta}(F_{ij})}~,
\end{equation}
yielding soft label vector $\wb_i \in \R^P$ for the $i$-th image proposals. 

By plugging in the
vMF probability density function of the common class and background
probability density function, the quotient $q_{\btheta}$ can be written as
\begin{equation}\label{eq:quotient_def}
  q_{\btheta}(\xb) \propto \exp \Big(\kappa\btheta^\top\xb - \log
u_{\bomega}^-(\xb)\Big)~.
\end{equation}
As shown in \cref{alg:coloc}, one can compute the soft labels via the softmax
operation, resembling the attention mechanism recently used for
MIL~\cite{ilse2018attention}. 

In the M-step, parameters $\btheta$ are updated by maximizing the surrogate expected log-likelihood using the posteriors computed in the E-step
\begin{equation}\label{eq:mstep_1_ap}
l(\btheta'; \btheta) = \sum_{i=1}^M \E_{p(\zb_i| F_i, \btheta)} \Big[\log
p(F_i|\zb_i, \btheta')\Big] = \sum_{i=1}^M \sum_{k=1}^P \wb_{ik} \log p(F_i|\zb_i=\eb_k, \btheta')~,
\end{equation}
where the weights $\wb_{ik}$ are computed in \cref{eq:e_step}. Lagrangian function is written as
\begin{equation}\label{eq:bag_dist_2}
    \LL(\btheta', \lambda) = l(\btheta'; \btheta) - \lambda (\norm{\btheta'}^2 - 1)
\end{equation}
By plugging in the log-likelihood term in \cref{eq:bag_dist_2} and computing the derivative w.r.t. $\btheta'$ and $\lambda$ we have
\begin{equation}\label{eq:mstep_2_ap}
\begin{split}
\nabla_{\btheta'}\LL(\btheta', \lambda) &= \sum_{i=1}^M \sum_{k=1}^P \wb_{ik} \nabla_{\btheta'}\log \Big(p_{\btheta'}^+(F_{ij})
\prod_{\substack{k=1\\k\neq j}}^P p_{\bomega}^-(F_{ik})\Big) - 2\lambda\btheta' = \kappa \sum_{i=1}^M \sum_{k=1}^P \wb_{ik} F_{ik}- 2\lambda\btheta'~. \\
\nabla_{\lambda}\LL(\btheta', \lambda) &= -\norm{\btheta'}^2 + 1.
\end{split}
\end{equation}
Finally, closed-form update rule 
\begin{equation}\label{eq:m_step_app}
\begin{split}
    \btheta &\gets \rm{norm}(\sum_{i=1}^M \tilde{\xb}_i)~, \\
    \text{where}\quad\tilde{\xb}_i &= \wb_i^\top F_i  =\sum_{k=1}^P \wb_{ik} F_{ik} ~,
\end{split}
\end{equation}
is derived by setting the derivatives to zero and solving for $\btheta'$ and $\lambda$. 

\subsection{Updating $\kappa$ in M-Step}
In the paper, $\kappa$ is a constant hyperparameter for all the novel classes and only $\btheta$ gets updated in the M-step. 
In this section, we propose a simple update rule for parameter $\kappa$ that can be used along \cref{eq:m_step_app} in the M-step. As shown in \cref{tab:coco60_coloc_kappa}, updating $\kappa$ with a our order-$0$ rule further improves our vMF-MIL COL results of the paper.

To find the optimal $\kappa$, we compute the derivative of the Lagrangian function in \cref{eq:bag_dist_2} w.r.t. $\kappa$
\begin{equation}\label{eq:mstep_kappa}
\begin{split}
\partial_{\kappa}\LL(\btheta', \lambda) &= \sum_{i=1}^M \sum_{k=1}^P \wb_{ik} \partial_{\kappa}\log \Big(p_{\btheta'}^+(F_{ij})
\prod_{\substack{k=1\\k\neq j}}^P p_{\bomega}^-(F_{ik})\Big) = \sum_{i=1}^M \sum_{k=1}^P \wb_{ik} \partial_{\kappa}\log p_{\btheta'}^+(F_{ik}) \\
&= \sum_{i=1}^M \sum_{k=1}^P \wb_{ik} \partial_{\kappa} (-\log Z(\kappa) + \kappa F_{ik}^\top \btheta') = -M \frac{\partial_{\kappa}Z(\kappa)}{Z(\kappa)} +  \rb^\top \btheta'~,
\end{split}
\end{equation}
where $\rb = \sum_{i=1}^M \sum_{k=1}^P \wb_{ik}\xb$ and $Z(\kappa)$ is vMF distribution normalization factor. A precise formula is known for $Z(\kappa)$, namely
\begin{equation}
\label{eq:exact-Z}
Z(\kappa) = \frac{(2\pi)^{d/2}\, I_{d/2-1}(\kappa)}{\kappa^{d/2-1}} ~.
\end{equation}
where $d$ is the feature dimension and $I$ is the modified Bessel function.
This formula is quoted in \cite{banerjee2005clustering}(2.2), but it is upside-down
compared to \cref{eq:exact-Z}, since we are defining $c_d(\kappa) = 1/Z(\kappa)$.

By plugging in $\btheta' = \rb/\norm{\rb}$ from \cref{eq:m_step_app} and setting the derivative to zero we get
\begin{equation}\label{eq:m_kappa}
    \frac{\partial_{\kappa}Z(\kappa)}{Z(\kappa)} = \frac{\norm{\rb}}{M} = \bar{r}~.
\end{equation}

\Cref{eq:m_kappa} is similar to what we see in vMF maximum-likelihood estimation, therefore, we can use the maximum-likelihood derivations from now on (See Appendix of \cite{banerjee2005clustering} equations A.7 to A.8) which leads to maximum-likelihood estimation
\begin{equation}\label{eq:exact_kappa}
    \kappa = A^{-1}_d(\bar{r})~,
\end{equation}
where $A_d$ is the ratio of Bessel functions,
\begin{equation}
    A_d(\kappa) = \frac{I_{d/2}(\kappa)}{I_{d/2-1}(\kappa)}~.
\end{equation}
%
It is still numerically difficult to compute the Bessel function in cases
where $d$ and $\kappa$ are large.  We are able to compute it
in python using the \verb:scipy.special.iv: function, only for values
of $d$ up to about $120$ and $\kappa$ up to about $700$.

\begin{figure}[t]
\centerline{
\includegraphics[height=2.5in]{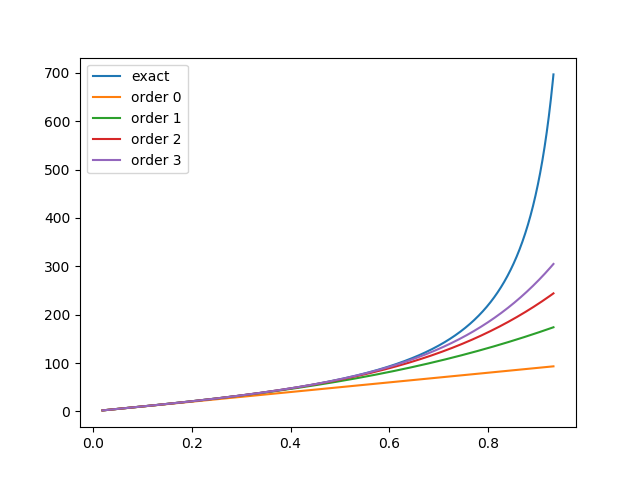}
}
\caption{\label{fig:orders}\small
\em Plot of different estimates of $\hat\kappa$ as a function of $\bar r$,
for dimension $d=100$.
At this resolution, the exact estimate is indistinguishable from the estimate 
\cref{eq:order-frac}.  The graph also shows approximations of different
orders, such as \cref{eq:order0} and \cref{eq:order1}, which are accurate for
small-to-medium values of $\bar r$, but not for larger values.  
However, the exact value of $\bar\kappa$ is extremely sensitive to
small variations in the value of $\bar r$, and it diverges to infinity
as $\bar r$ approaches $1$.  For this reason it may not be good practice
(as is
verified by our experiments) to use the exact estimate of $\bar\kappa$
in clustering.
}
\end{figure}
To address this difficulty, different formulae are given for estimating the optimal value
of $\hat\kappa$.
\begin{enumerate}
\item 
A formula due to Mardia and Jupp~\cite{mardia2009directional} is given in 
\cite{banerjee2005clustering}(4.2)
as
\[
\hat\kappa \approx d \bar r \left(1 + \frac{d}{d+2} \bar r^2 + \frac{d^2(d+8)}{(d+2)^2(d+4)}\bar r^4 \right)~.
\]
For large values of $d$, this is almost the same as
\begin{equation}
\label{eq:order-2}
\hat\kappa \approx d \bar r \left(1 + \bar r^2 + \bar r^4 \right) ~.
\end{equation}
\item 
Banerjee \etal \cite{banerjee2005clustering} derive an estimate (unnumbered, above \cite{banerjee2005clustering} (4.4)),
\begin{equation}
\label{eq:order-frac}
\hat\kappa \approx \frac{d \bar r}{1 - \bar r^2}
\end{equation}
which has series expansion
\begin{equation}
\label{eq:order-infinity}
\hat\kappa \approx d \bar r (1 + \bar r^2 + \bar r^4 + \ldots ) ~.
\end{equation}
Since $0 \le \bar r < 1$, this series will converge, albeit slowly
for $\bar r$ close to $1$.
Thus, it is seen that \cref{eq:order-2} is simply a truncated approximation
to the infinite series \cref{eq:order-infinity}.
We shall refer to \cref{eq:order-2} (perhaps somewhat inexactly) as the
``order-$2$'' approximation to \cref{eq:order-infinity}, since the approximation
to $1 / (1 - \bar r^2)$ contains terms up to second order in $\bar r^2$.
\item
It is also possible to consider approximations of other orders for
\cref{eq:order-frac}, including in particular the $0$-order approximation
\begin{equation}
\label{eq:order0}
\hat\kappa \approx d \bar r ~,
\end{equation}
first order approximation
\begin{equation}
\label{eq:order1}
\hat\kappa \approx d \bar r \left(1 + \bar r^2 \right) ~,
\end{equation}
and the third-order approximation.
\item Another empirically derived formula is also given in 
\cite{banerjee2005clustering}(4.4).  However, we observe (see \cref{fig:orders})
that the approximation
\cref{eq:order-frac} is already a very close approximation, and the use of
\cite{banerjee2005clustering}(4.4) is not warranted.
\end{enumerate}
We show graphs of the approximations of $\hat\kappa$ for various approximation,
and the exact solution in \cref{fig:orders}.

\paragraph{Results}
As pointed out in the caption to \cref{fig:orders}, the exact estimate
of $\hat\kappa$ may not be a good choice for clustering, in the case
where $\bar r$ approaches $1$ (meaning that the data has small spread).

In \cref{tab:coco60_coloc_kappa}, we try all the different formulas described above
to estimate a value of $\hat\kappa$.  We also report the results in
the main paper where $\kappa$ is kept constant. The experiments 
show the following outcomes.
\begin{enumerate}
\item Order-$\infty$ in \cref{eq:order-frac} performs notably worse, presumably because of
the sensitivity to the value of $\bar r$, which is computed from a
relatively small number of samples in our few-shot learning scenario.
\item The order-$1$ to order-$3$ estimates perform approximately the
same as fixing $\hat\kappa$.
\item The order-$0$ approximation, $\hat\kappa = d \bar r$, gives the best
results.  In our COL experiments, $d = 512$, so setting
$\hat\kappa = d\bar r$ places an upper bound of $512$ on the value
of $\hat\kappa$.
\end{enumerate}

\begin{table}[t]
\caption{\em CorLoc(\%) and mAP(\%) results with $\kappa$ estimations for the task of COL on novel object classes on the COCO60 dataset with support set size $K=5$ and $K=10$.}
\label{tab:coco60_coloc_kappa}
\vspace{0.1cm}
\centering
\resizebox{0.65\linewidth}{!}{%
\begin{tabular}{l||cc|cc}
\toprule
\multicolumn{1}{c}{\multirow{2}{*}{$\hat\kappa$}} & \multicolumn{2}{c}{K = 5}   & \multicolumn{2}{c}{K=10} \\
\multicolumn{1}{c}{} & \multicolumn{1}{c}{CorLoc@0.5} & \multicolumn{1}{c}{mAP@0.5} & \multicolumn{1}{c}{CorLoc@0.5} & \multicolumn{1}{c}{mAP@0.5} \\
\midrule
Constant  & $34.8$ & $18.6$ & $36.9$ & $20.0$ \\
$d \bar r (1 + \bar r^2 + \bar r^4 + \ldots )$  & $20.1$ & $11.3$ & $24.6$ & $13.7$ \\
$d \bar r (1 + \bar r^2 + \bar r^4 + \bar r^6)$  & $31.8$ & $17.7$ & $34.7$ & $19.0$ \\
$d \bar r (1 + \bar r^2 + \bar r^4)$  & $33.1$ & $18.6$ & $36.3$ & $19.5$ \\
$d \bar r (1 + \bar r^2)$  & $34.7$ & $19.0$ & $37.5$ & $19.9$ \\
$d \bar r$  & ${\bf 35.7}$ & ${\bf 19.6}$ & ${\bf 38.2}$ & ${\bf 20.2}$ \\
\bottomrule 
\end{tabular}
}
\end{table}

\section{Modeling with Gaussian Distribution}\label{sec:gaussian}
In \cref{sec:ablation}, we conduct experiment with Gaussian distribution used to model the common object distribution. We assume the common object distribution is Gaussian with mean $\btheta$ and diagonal covariance matrix $\sigma^2 I$, i.e., $\xb \sim \NN(\btheta, \sigma I)$
\begin{equation}
    p^+_{\btheta_c}(\xb) = \frac{1}{Z} \exp \left(-\frac{\norm{\xb - \btheta}^2}{2\sigma^2}\right)~,
\end{equation}
and plug the distribution into \cref{eq:e_step_app} to get the soft label update rule as
\begin{equation}\label{eq:e_step_gaussian_app}
\wb_{ik} = \frac{\exp \Big( -\frac{\norm{F_{ik}-\btheta}^2}{2\sigma^2} - \log
u_{\bomega}^-(F_{ik})\Big)}{\sum_{j=1}^P \exp \Big(-\frac{\norm{F_{ij}-\btheta}^2}{2\sigma^2} - \log
u_{\bomega}^-(F_{ij})\Big)}~.
\end{equation}
E-step is computed by setting the derivative of \cref{eq:mstep_1_ap} w.r.t. $\btheta$ to zero
\begin{equation}\label{eq:m_step_gaussian_app}
    \btheta \gets \frac{1}{M} \sum_{i=1}^M\sum_{k=1}^P \wb_{ik} F_{ik} =
\frac{1}{M}\sum_{i=1}^M \wb_i^\top F_i~,
\end{equation}
\section{MI-SVM WSOD Baseline}\label{sec:mi-svm}
To the best of our knowledge there is no WSOD algorithm for few-shot setting in
the literature. However, WSOD with knowledge transfer
methods~\cite{rahimi2020pairwise, uijlings2018revisiting,hoffman2016large,deselaers2010localizing} 
are closely related to our work. In this section, we
describe a slightly modified version of~\cite{uijlings2018revisiting} and
discuss its differences to the proposed method. In \cref{sec:experiments}, we
empirically compare our work against~\cite{uijlings2018revisiting}.

Similar to our approach, each image is represented as a bag of bounding box
proposals $B$ and their features $F$.
Learning is performed on one target class
$c \in \LL$ at a time. Similar to our WSOD approach, the support set is split into positive images
which has the target class and a negative set of
images without the target class. Then, a linear SVM appearance model is employed
to iteratively learn class $c$ by alternating between two steps:
\begin{itemize}[leftmargin=*]
    \item {\bf Re-training:} Train a binary SVM given the currently selected
proposals from the positive images and the proposals in negative images.
    \item {\bf Re-localization:} Given the current SVM select the proposal with
the highest score from each positive image. 
In~\cite{uijlings2018revisiting},
the re-localization is guided by a class-agnostic objectness measure to guide
the selection toward objects. 
Therefore, the selection for a positive image
$\Ib$ with bounding box proposals $B$ is updated as
    \begin{equation}
        b^* = \argmax_{b\in B}\quad \text{SVM}(\Ib, b) + \gamma O(\Ib, b)~,
    \end{equation}
    where $O$ is the objectness model and $\gamma$ adjusts its importance.
 \end{itemize}
The algorithm is initialized with complete image bounding box proposal and
alternates between above steps until convergence. 
We use highly efficient GPU
solver in~\cite{amos2017optnet} for SVM optimization. 
Finally, test proposal
$\xb$ from the test set $\DD_{\rm test}$ is scored using the SVM trained for
each class. 
To have a fair comparison, we use the same Faster-RCNN model trained
on the base classes as we used in our model to extract bounding box and feature
proposals from all images. 
For the objectness model $O$, we learn a
class-agnostic logistic regression model on the extracted feature. 
We employ the hard negative mining in~\cite{uijlings2018revisiting} to improve the
performance of the classifier. 
In this approach the negative set is initialized
with full negative image features and the hardest negative proposal within each
image is added to the negative set after each re-training step.

The expectation and maximization steps in our method are analogous to
re-localization and re-training steps in~\cite{uijlings2018revisiting}. 
In MI-SVM, only the proposal with the highest score is
labeled positive in the re-localization step while our COL method infers soft labels
in the expectation step via an attention mechanism. Using soft labels could be
beneficial as they reflect the uncertainty in choosing the common object. 


\section{Qualitative Results}
\begin{figure}[h]
\centering
\includegraphics[width=0.6\linewidth]{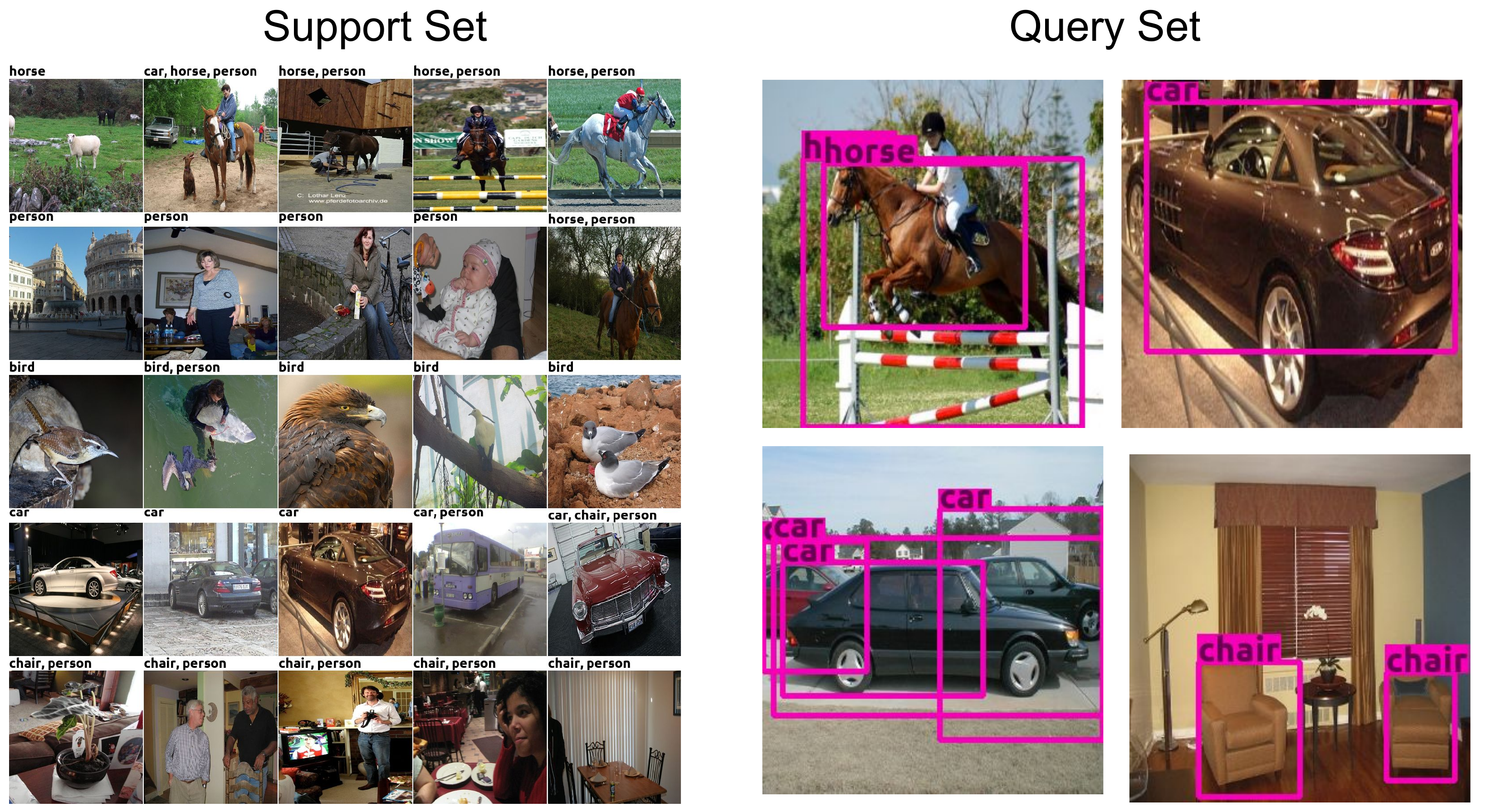}
\caption{\em\label{fig:wsod_pascal_qual} Few-shot WSOD on PASCAL VOC with $N=K=5$. Given the support set shown on the left side the algorithm detects the object on the $4$ different query images on the right side. The algorithm fails to detect person in the first query image but successfully detects other target objects.}
\end{figure}

\begin{figure}[h]
\centering
\includegraphics[width=0.6\linewidth]{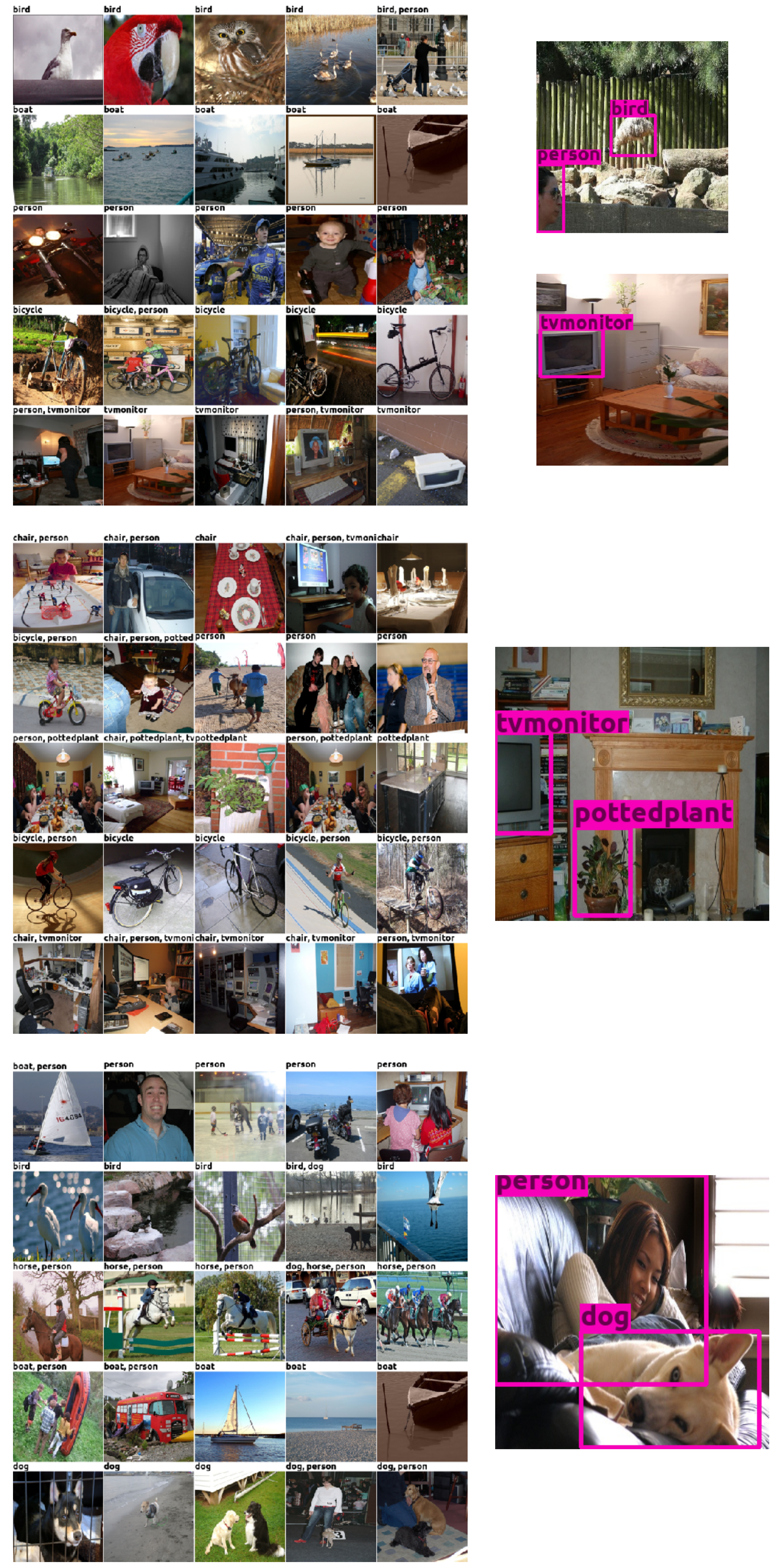}
\caption{\em\label{fig:wsod_pascal_qual_1} Few-shot WSOD on PASCAL VOC with $N=K=5$. Given the support set shown on the left side the algorithm detects the object in query images on the right side.}
\end{figure}

\begin{figure}[h]
\centering
\includegraphics[width=0.6\linewidth]{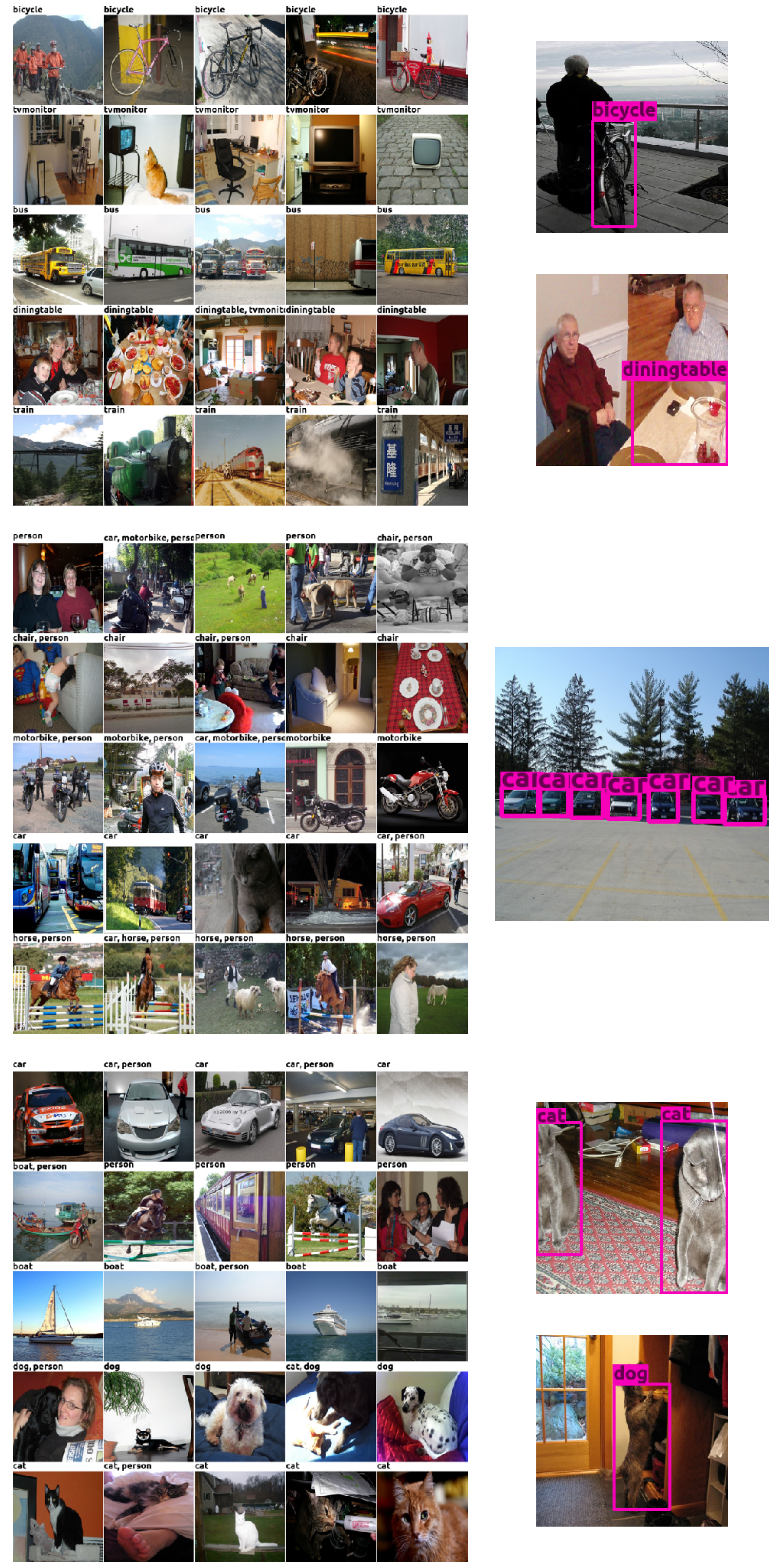}
\caption{\em\label{fig:wsod_pascal_qual_2} Few-shot WSOD on PASCAL VOC with $N=K=5$. Given the support set shown on the left side the algorithm detects the object in query images on the right side.}
\end{figure}

\begin{figure}[h]
\centering
\includegraphics[width=0.6\linewidth]{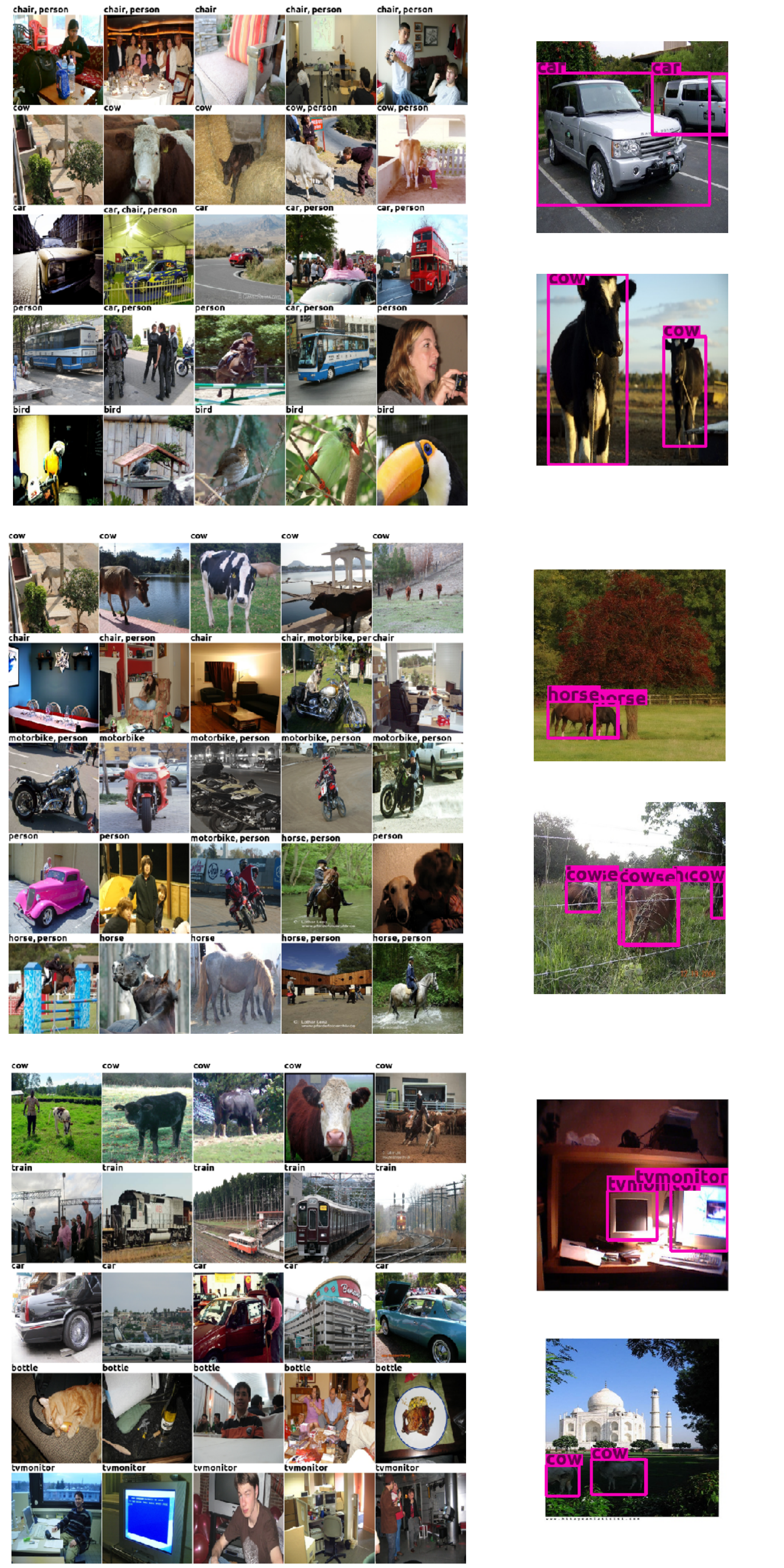}
\caption{\em\label{fig:wsod_pascal_qual_3} Few-shot WSOD on PASCAL VOC with $N=K=5$. Given the support set shown on the left side the algorithm detects the object in query images on the right side.}
\end{figure}

\begin{figure}[h]
\centering
\includegraphics[width=0.6\linewidth]{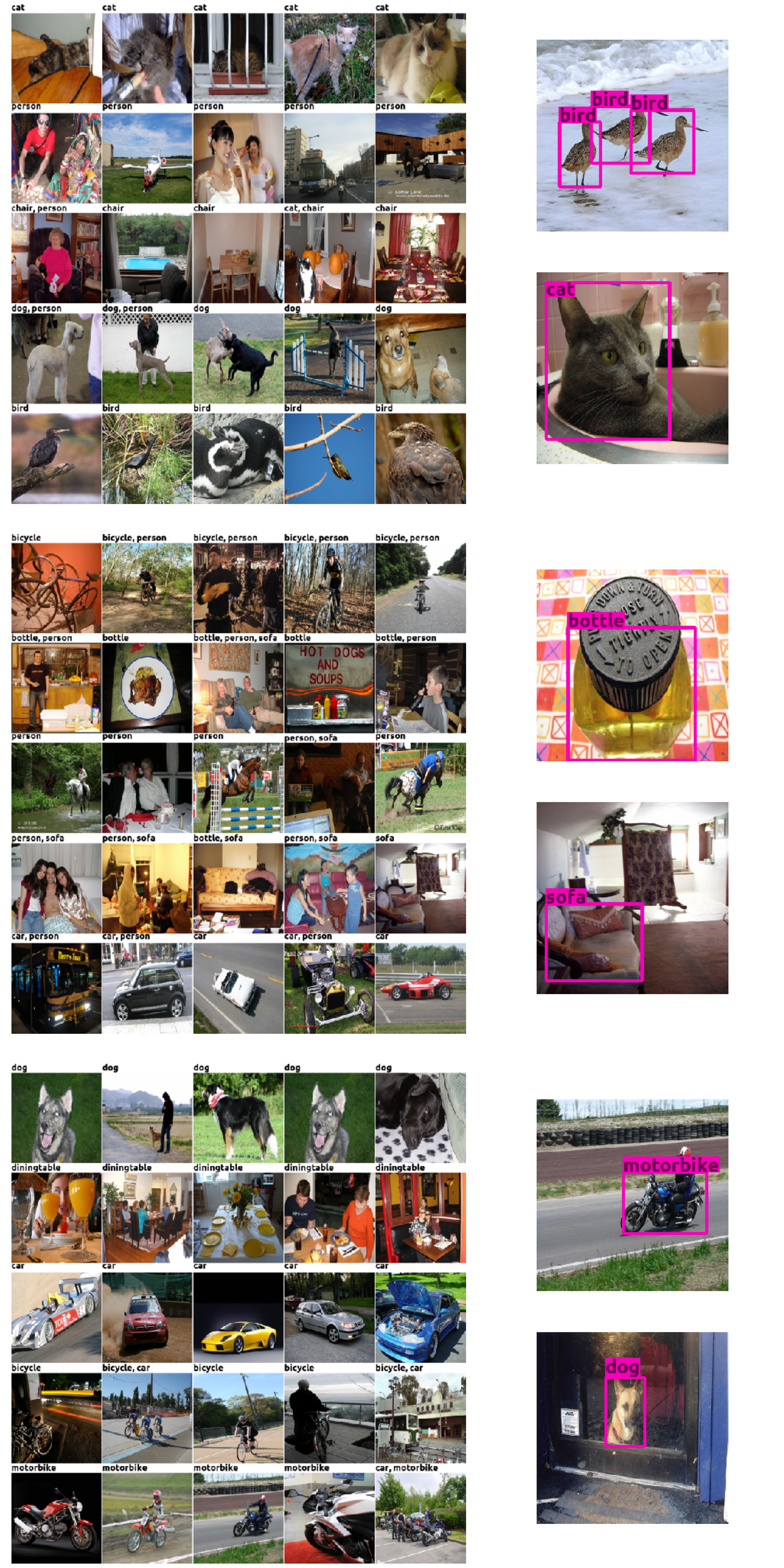}
\caption{\em\label{fig:wsod_pascal_qual_4} Few-shot WSOD on PASCAL VOC with $N=K=5$. Given the support set shown on the left side the algorithm detects the object in query images on the right side.}
\end{figure}

\begin{figure}[h]
\centering
\includegraphics[width=0.6\linewidth]{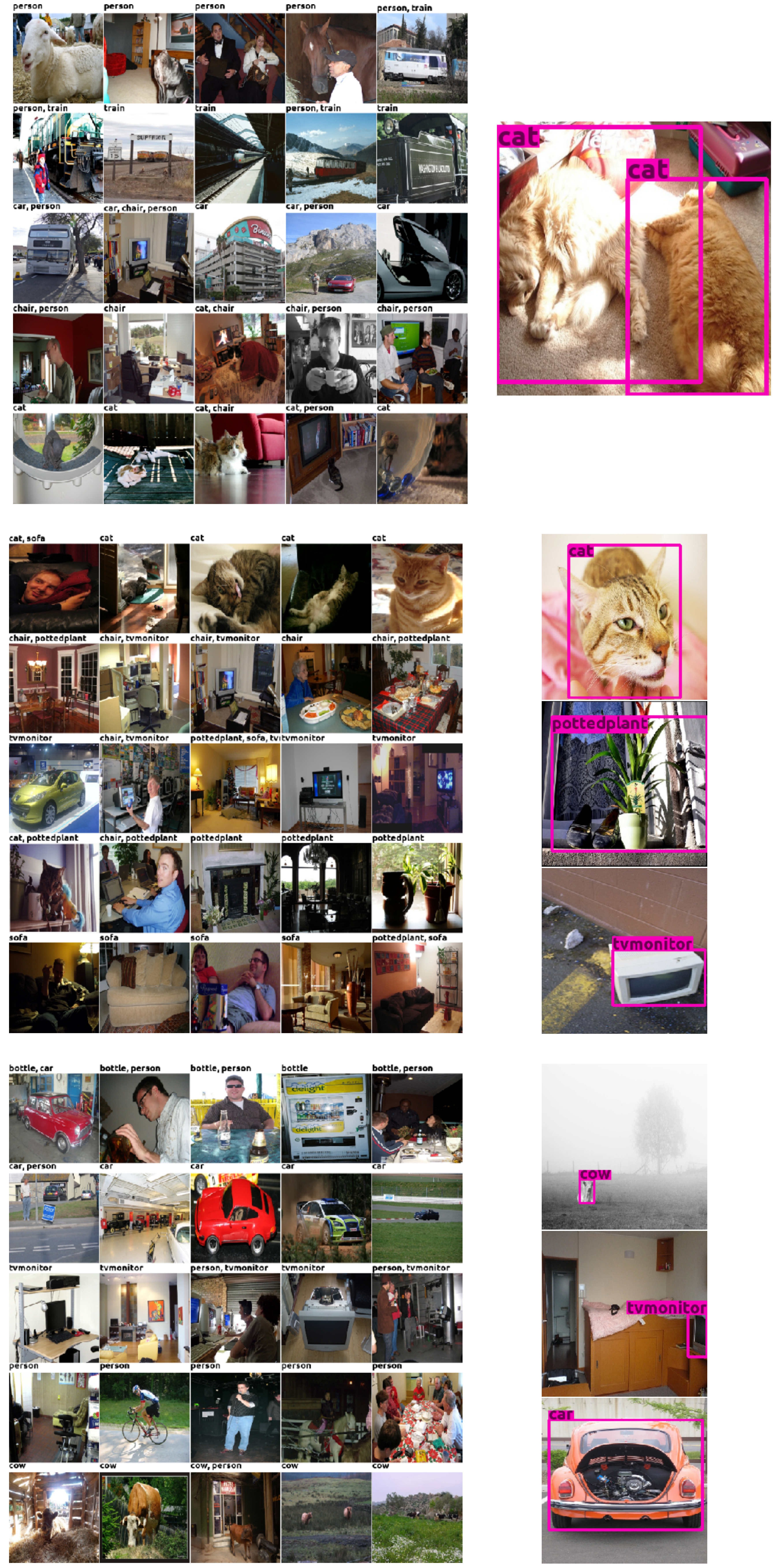}
\caption{\em\label{fig:wsod_pascal_qual_5} Few-shot WSOD on PASCAL VOC with $N=K=5$. Given the support set shown on the left side the algorithm detects the object in query images on the right side.}
\end{figure}

We show some of the success cases of the experiments in the paper. Our first example in \cref{fig:wsod_pascal_qual} shows vMF-MIL performance on a single few-shot WSOD problem. Given the support set with only image-level annotations the algorithm learns to detect the target objects in the query set. We sample $4$ query images to evaluate the algorithm performance in detecting different target objects. Except {\tt person} in the first query image, vMF-MIL successfully detects other target objects. More few-shot WSOD tasks are shown in \cref{fig:wsod_pascal_qual_1}, \ref{fig:wsod_pascal_qual_2}, \ref{fig:wsod_pascal_qual_3}, \ref{fig:wsod_pascal_qual_4} and \ref{fig:wsod_pascal_qual_5}.

\begin{figure}[h]
\centering
\includegraphics[width=0.7\linewidth]{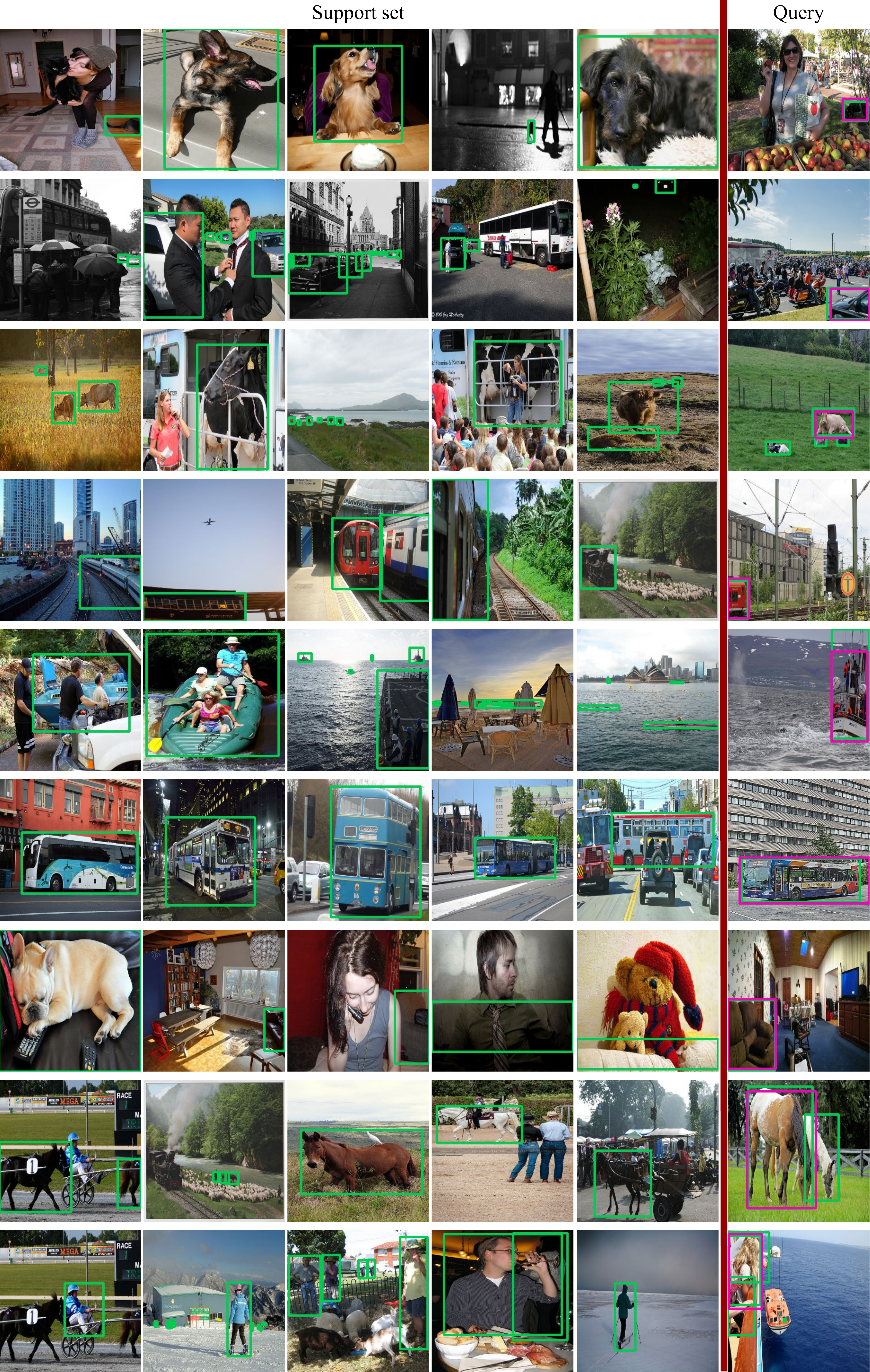}
\caption{\em\label{fig:coloc_coco_qual} 5-shot common object localization ($N=1$) on MS COCO. Each row shows one common object localization problem. Ground-truth annotations (shown in green) are just for visualization and are not used in the algorithm. Top query bounding box prediction for each problem is shown in pink.}
\end{figure}

Finally, \cref{fig:coloc_coco_qual} shows some of the success cases in localizing the target object in the query image for the task of common object localization in \cref{sec:coloc}. All the target objects ({\tt dog, car, cow, train, boat, bus, sofa, horse, person}) shown in this figure are novel. Also, ground-truth annotations are only shown for better visualization and are not used in learning. 

\end{document}